%% file: main.tex
\documentclass[10pt,twocolumn,letterpaper]{article}

\usepackage{cvpr}              

\input{preamble}

\definecolor{cvprblue}{rgb}{0.21,0.49,0.74}
\usepackage[pagebackref,breaklinks,colorlinks,allcolors=cvprblue]{hyperref}

\def\paperID{*****} 
\def\confName{CVPR}
\def\confYear{2025}

\title{Why We Feel:~Breaking Boundaries in Emotional Reasoning with Multimodal Large Language Models}

\author{
{\bf Yuxiang Lin}\textsuperscript{1}\thanks{Work done during internship at UW.} \quad
{\bf Jingdong Sun}\textsuperscript{2}\footnotemark[1] \quad
{\bf Zhi-Qi Cheng}\textsuperscript{3}\thanks{Corresponding author~(zhiqics@uw.edu).} \quad
{\bf Jue Wang}\textsuperscript{4}\footnotemark[1] \quad
{\bf Haomin Liang}\textsuperscript{5}\footnotemark[1]\\
{\bf Zebang Cheng}\textsuperscript{5}\footnotemark[1] \quad
{\bf Yifei Dong}\textsuperscript{3}\quad
{\bf Jun-Yan He}\textsuperscript{6} \quad
{\bf Xiaojiang Peng}\textsuperscript{5} \quad
{\bf Xian-Sheng Hua}\textsuperscript{7}\\[6pt]
\textsuperscript{1}Georgia Institute of Technology \quad
\textsuperscript{2}Carnegie Mellon University \quad
\textsuperscript{3}University of Washington \\
\textsuperscript{4}Shenzhen Institute of Advanced Technology \quad
\textsuperscript{5}Shenzhen Technology University \\
\textsuperscript{6}Alibaba Group \quad
\textsuperscript{7}Tongji University
}

\begin{document}
\maketitle
\input{sections/abstract}

\input{sections/introduction}

\input{sections/related_work}

\input{sections/problem}

\input{sections/benchmark}

\input{sections/experiment}
\input{sections/conclusion}

\clearpage
{
    \small
    \bibliographystyle{ieeenat_fullname}
    \bibliography{main}
}

\clearpage
\appendix
\input{Supplementary/dataset_and_code_access}
\input{Supplementary/baseline_models}
\input{Supplementary/bench_details}



\end{document}

%% file: preamble.tex
\usepackage{xcolor}         
\usepackage{soul}
\usepackage{color}
\usepackage{colortbl}
\usepackage{pifont}
\usepackage{enumitem}
\usepackage{amsmath}
\usepackage{graphicx}
\usepackage{amsthm}
\usepackage{algorithmic}
\usepackage{algorithm}
\usepackage{wrapfig}
\usepackage{dsfont}
\usepackage{multirow}
\usepackage{siunitx}

\usepackage{soul}

\definecolor{lightOrange}{rgb}{1.0, 0.93, 0.75}  
\sethlcolor{lightOrange}                          

\definecolor{LightRed}{RGB}{255,204,204}
\definecolor{LightBrown}{RGB}{222,184,135}
\definecolor{LightYellow}{RGB}{255,255,204}
\definecolor{LightGreen}{RGB}{144,238,144}
\definecolor{lightOrange}{RGB}{255, 223, 186}
\newcommand{\xmark}{\ding{55}}
\newcommand{\cmark}{\ding{51}}

%% file: sections/abstract.tex
\begin{abstract}
Most existing emotion analysis emphasizes \emph{which} emotion arises (e.g., happy, sad, angry) but neglects the deeper \emph{why}. We propose \textbf{Emotion Interpretation (EI)}, focusing on \emph{causal factors}—whether explicit (e.g., observable objects, interpersonal interactions) or implicit (e.g., cultural context, off-screen events)—that drive emotional responses. Unlike traditional emotion recognition, EI tasks require \emph{reasoning about triggers} instead of mere labeling. To facilitate EI research, we present \textbf{EIBench}, a large-scale benchmark encompassing \num{1615} \emph{basic} EI samples and \num{50} \emph{complex} EI samples featuring multifaceted emotions. Each instance demands rationale-based explanations rather than straightforward categorization. We further propose a \emph{Coarse-to-Fine Self-Ask (CFSA)} annotation pipeline, which guides Vision-Language Models (VLLMs) through iterative question-answer rounds to yield high-quality labels at scale. Extensive evaluations on open-source and proprietary large language models under four experimental settings reveal consistent performance gaps—especially for more intricate scenarios—underscoring EI's potential to enrich empathetic, context-aware AI applications. Our benchmark and methods are publicly available at \href{https://github.com/Lum1104/EIBench}{https://github.com/Lum1104/EIBench}, offering a foundation for advanced multimodal causal analysis and next-generation affective computing.
\end{abstract}

%% file: sections/introduction.tex
\begin{figure*}[t]
\centering
\includegraphics[width=0.98\textwidth]{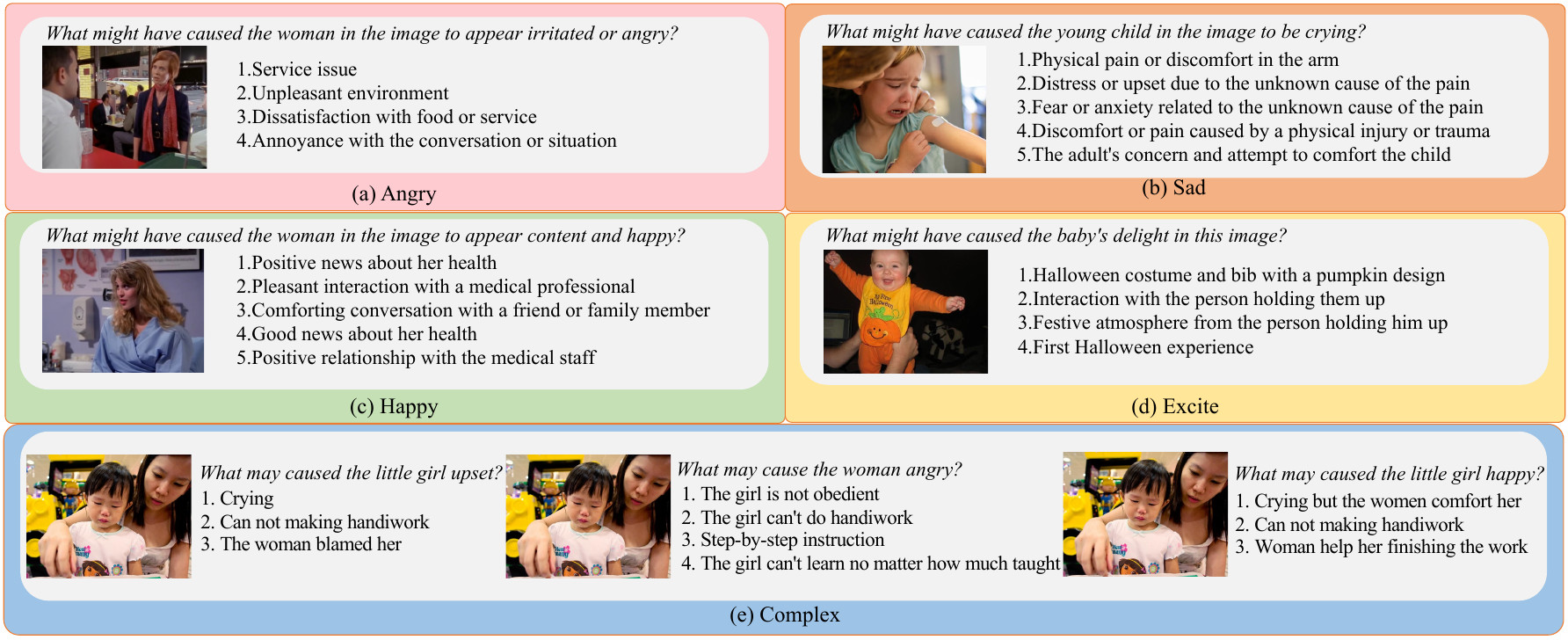}
\vspace{-2mm}
\caption{\small
Illustrative examples of \emph{Emotion Interpretation} in five categories: 
\textbf{(a)}~Angry, \textbf{(b)}~Sad, \textbf{(c)}~Happy, \textbf{(d)}~Excited, and \textbf{(e)}~Complex. 
Each panel shows a scenario with potential triggers (e.g., service frustrations, medical news, festive attire, family interactions). 
In (e), multiple triggers or viewpoints co-occur: a child upset about craft-making and a caregiver’s frustration. 
By integrating facial cues, context, and domain knowledge, this approach surpasses mere emotion labeling, clarifying \emph{why} individuals feel a certain way.
}
\label{trigger_shows}
\vspace{-4mm}
\end{figure*}

\section{Introduction}
\label{sec:intro}
Emotion analysis plays a pivotal role in diverse fields such as \emph{human-computer interaction} (HCI)~\citep{JAIN2023102662,ma2022should,10.1145/3313831.3376187,yang2019understanding}, \emph{healthcare}~\citep{dahl2007sleep,saarni2007emotional,tronick2018emotions}, and \emph{market research}~\citep{cambria2017affective,caruelle2022affective,srivastava2024modern}. While recent advances in \emph{emotion recognition} (e.g., predicting whether someone feels “happy” or “sad”) have offered valuable insights, they often overlook the deeper question of \emph{why} a particular emotion arises. Because emotions can be subtle and highly subjective, merely labeling the emotional state fails to capture the nuanced triggers that might underlie or amplify the expressed affect.

To address the limitations of focusing on \emph{which} emotion is present, we highlight the significance of \emph{emotion interpretation}, where the objective is to explain \emph{why} an individual experiences a specific emotional response. In practical applications (e.g., empathic virtual assistants, mental health counseling, user experience evaluations), identifying the emotion alone provides incomplete information if underlying triggers remain unknown. For instance, knowing a user is “angry” but not understanding whether the anger stems from waiting in a queue, receiving unfavorable feedback, or personal stressors hampers targeted interventions. Consequently, there is a need for systematic frameworks to help AI models identify and communicate reasons behind emotional states, thereby enabling more empathetic and context-aware intelligent services.

In response, we propose \textbf{Emotion Interpretation (EI)}, shifting emphasis from \emph{recognizing} an emotion label to \emph{reasoning about} triggers behind it. Unlike classical emotion recognition, EI centers on \emph{why} the emotional state arises and accommodates both explicit cues (e.g., visible objects, interpersonal interactions) and implicit or off-screen factors (e.g., historical context, hidden storylines). As shown in Figure~\ref{trigger_shows}, EI spans scenarios from straightforward triggers (e.g., prolonged waiting leading to frustration) to complex ones with multiple emotional facets (e.g., overlapping sadness and resentment). Modern \emph{Vision-Language Models (VLLMs)}~\citep{qwen,minigpt4v2,vllava,llava,wang2023skeleton,liu2024llava,liu2023improved,li2023otter} hold promise for EI by integrating visual cues with rich world knowledge to produce explanatory text.

Despite progress in multimodal learning, most existing datasets still focus on \emph{emotion classification} rather than \emph{causal factors}. Moreover, standard unimodal benchmarks seldom capture how vision, language, and context interact to explain emotional triggers. To address this gap, we create the \emph{EIBench} dataset, comprising \num{1615} well-annotated \emph{basic} EI samples plus \num{50} \emph{complex} EI samples. Each sample challenges models to reason more deeply about multi-layered or co-occurring emotions. This dataset thus supports advanced evaluation protocols reflecting real-world complexity, in line with the push for more sophisticated multimodal benchmarking. Building on these objectives, our main contributions include:
\begin{enumerate}
    \item \textbf{Task Definition:} We formally define \emph{Emotion Interpretation (EI)} as moving beyond simple emotion labeling toward revealing the \emph{causes} behind an individual’s emotional state. This shift enables more empathetic and context-aware AI systems.
    \item \textbf{Benchmark Dataset:} We introduce \textbf{EIBench}, a large-scale resource specifically aimed at EI, spanning four primary emotion categories (e.g., angry, sad, excited, happy) and \emph{complex} scenarios where multiple emotions interlace. This dataset allows for evaluating diverse dimensions of emotional interpretation.
    \item \textbf{Annotation Method (CFSA):} We develop a \emph{Coarse-to-Fine Self-Ask} (CFSA) procedure inspired by \emph{chain-of-thought} reasoning~\citep{self-ask,self-refine,tree-of-thought,graph-of-thoughts,mm-cot,auto-cot}. By leveraging advanced Vision-Language Models in a semi-automated workflow, CFSA collects and refines multi-round insights about emotional triggers, yielding high-quality annotations that capture both explicit and implicit factors.
    \item \textbf{Comprehensive Evaluation:} We perform systematic experiments on both open-source and proprietary LLMs under four different testing settings (e.g., using image captions, chain-of-thought prompting, and persona-based variations). Our results highlight significant performance gaps across these models. Notably, some proprietary models (e.g., Claude-3, ChatGPT-4) excel in simpler emotion interpretation tasks yet struggle to maintain the same level of accuracy in multi-perspective, complex scenarios—indicating the need for enhanced interpretative strategies.
\end{enumerate}
\vspace{2mm}


%% file: sections/related_work.tex
\begin{table*}[!th]
\begin{minipage}{1\textwidth}
\centering
\vspace{-4mm}
\caption{\small This table demonstrates how the CFSA Method interprets excitement and joy at an LGBT event, where \sethlcolor{pink}\hl{pink} text highlights generated captions, \sethlcolor{yellow}\hl{yellow} text shows user query content, and \sethlcolor{lightOrange}\hl{light orange} text corresponds to matched triggers.}
\vspace{-1mm}
\scalebox{0.65}{
    \begin{tabular}{l|lp{9.5cm} }
\toprule
\multicolumn{3}{l}{\bf An Example of Comprehending Excitement} \\
\midrule
User Question & \multicolumn{2}{c}{\includegraphics[height=4.5cm]{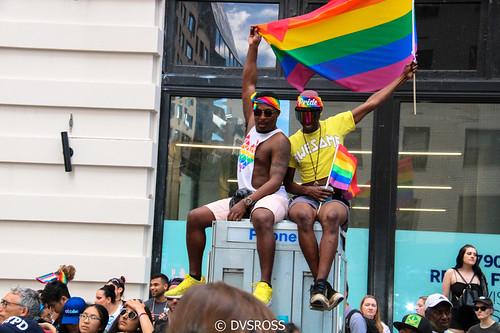}} \\
{\small \textcolor{orange}{(VLLM)}} & \multicolumn{2}{l}{\textit{What might have caused \textbf{the man holding the rainbow flag} to \textbf{become so excited and joyful}?}} \\
\midrule
User Question Preprocessing & Question Parse & Who is the person user talks about? \\
{\small \textcolor{orange}{(VLLM)}} & & The man \sethlcolor{pink}\hl{holding the rain bow flag}. \\
& & What is the user's demand? \\
& & To know why the man is so \sethlcolor{yellow}\hl{excited and joyful}. \\
& Question Reconstruction & \sethlcolor{pink}\hl{The man holding the rainbow flag is a proud and enthusiastic participant in a parade, possibly celebrating LGBTQ+ pride or a similar event. He is waving the flag high in the air, showing his support and joy.} \sethlcolor{yellow}\hl{The user wants to know why the man is so excited and joyful.} \\
\midrule
General Self-Ask Thinking & \multicolumn{2}{p{13.5cm}}{1. How many people in this image, what are they doing?} \\
{\small \textcolor{orange}{(Question: Text-only LLM (GPT-4))}}& \multicolumn{2}{p{13.5cm}}{... They are \sethlcolor{lightOrange}\hl{standing on top of a box} ... are likely participating in a \sethlcolor{lightOrange}\hl{parade or public event}.} \\
{\small \textcolor{orange}{(Answer: VLLM)}}& \multicolumn{2}{p{13.5cm}}{2. What does ``the person user mentioned'' do in this image? Does he have interaction with others?} \\
& \multicolumn{2}{p{13.5cm}}{... is \sethlcolor{lightOrange}\hl{holding the rainbow flag and waving it high in the air}...}\\
& \multicolumn{2}{p{13.5cm}}{3. What is the facial expression of the person mentioned?}\\
& \multicolumn{2}{p{13.5cm}} {The facial expression of the person holding the flag is one of excitement and joy.}\\
& \multicolumn{2}{p{13.5cm}}{4. How about his body language?}\\
& \multicolumn{2}{p{13.5cm}}{... holding the flag is \sethlcolor{lightOrange}\hl{expressive and energetic}, ... \sethlcolor{lightOrange}\hl{active participation in the event}.}\\
\midrule
Scenario Specific Self-Ask Thinking & \multicolumn{2}{p{13.5cm}}{1. What specific ... the man holding the rainbow flag ... relates to his excitement and joy?} \\
{\small \textcolor{orange}{(VLLM)}} & \multicolumn{2}{p{13.5cm}}{The man is \sethlcolor{lightOrange}\hl{celebrating LGBTQ+ pride} ... as indicated by the rainbow flag. ... participating in such an event is a way for him to express his \sethlcolor{lightOrange}\hl{support for the LGBTQ+ community and celebrate their achievements and progress}.} \\
& \multicolumn{2}{p{13.5cm}}{2. What factors, such as personal experiences or beliefs, might contribute to the man's enthusiastic participation in the parade or event?} \\
& \multicolumn{2}{p{13.5cm}}{... he may have \sethlcolor{lightOrange}\hl{friends or family members who identify as LGBTQ+} ...}\\
& \multicolumn{2}{p{13.5cm}}{3. How does the man's enthusiasm and joy impact the overall atmosphere and energy of the crowd and the event?}\\
& \multicolumn{2}{p{13.5cm}} {The man's enthusiasm ... \sethlcolor{lightOrange}\hl{energy of the crowd and the event}...}\\
& \multicolumn{2}{p{13.5cm}}{4. Are there any interactions or reactions from the crowd that might be influencing or being influenced by the man's excitement and joy?}\\
& \multicolumn{2}{p{13.5cm}}{... provide him with \sethlcolor{lightOrange}\hl{a sense of validation and encouragement} for his participation in the event.}\\
\midrule
Emotion Summarization &  \multicolumn{2}{p{13.5cm}}{1. Celebrating LGBTQ+ pride or a similar event. 2. Supporting the LGBTQ+ community.}\\
(Emotional Triggers) & \multicolumn{2}{p{13.5cm}}{3. Expressing his personal beliefs and values. 4. Feeling a sense of unity and belonging with the crowd. 5. Being part of a positive and uplifting event. 6. Standing on top of a box.}\\
\bottomrule
\end{tabular}
}
\vspace{-4mm}
\label{tab:ec_example}
\end{minipage}
\end{table*}

\section{Related Work}
\label{sec:related_work}
We review the most relevant lines of research that inform our work on \emph{Emotion Interpretation (EI)}. Unlike prior methods that primarily \emph{recognize} an emotion label, our approach aims to \emph{interpret} the latent triggers behind that emotion.

\subsection{Context-Aware Emotion Recognition}
\label{subsec:caer_related_work}
\emph{Facial Expression Recognition} (FER) focuses on perceiving emotion from faces alone~\citep{ran,scn,psr,poster,poster++,fer-former,expMAE}, whereas \emph{Context-Aware Emotion Recognition} (CAER) leverages broader contextual cues~\citep{emotic,context-de-confounded,vllmcaer,bhattacharya2020step,ruan2020context,mittal2020emoticon,li2021human,heco,zhang2019context} such as body language or background details. For instance, EMOTIC~\citep{emotic} integrates the body region and the global scene, while CAER-S~\citep{caer-s} captures human social contexts from movie clips. Recently, \citet{vllmcaer} exploited \emph{commonsense knowledge} from Vision-Language Models (VLLMs) to boost CAER performance. However, these endeavors predominantly concentrate on determining \emph{which} emotion is expressed, not on uncovering \emph{why} the emotion arises.

\subsection{Emotion Recognition with LLMs}
\label{subsec:er_llm_related_work}
The advent of Large Language Models (LLMs) has introduced new possibilities for \emph{explainable} emotion recognition~\citep{cheng2024sztu,emotionllama,commonsense,implicit-sentiment,lei2023instructerc,emovit}. Some approaches use chain-of-thought prompting to help LLMs identify hidden or implicit sentiments~\citep{implicit-sentiment}, whereas others employ retrieval-augmented pipelines for conversational emotion detection~\citep{lei2023instructerc}. In the multimodal domain, VLLMs~\citep{llava,vllava,liu2024llava} enable image-grounded reasoning~\citep{cheng2024mips,emovit,vllmcaer}, but these systems still center on labeling emotions rather than interpreting the underlying \emph{causes}. By contrast, EI explores deeper triggers—even those not directly visible—and generates generative, flexible explanations.

\begin{table*}[t]
\centering
\scriptsize
\caption{\small 
A structured comparison of six major emotion-related tasks, highlighting their objectives and formal input--output relationships. 
\textbf{FER} = \emph{Facial Emotion Recognition}, 
\textbf{CAER} = \emph{Context-Aware Emotion Recognition}, 
\textbf{ER with LLMs} = \emph{Emotion Recognition with Large Language Models}, 
\textbf{HS} = \emph{Humor Study}, 
\textbf{ECE} = \emph{Emotion Cause Extraction}, 
\textbf{EI} = \emph{Emotion Interpretation}.
}
\label{tab:related_work}
\vspace{-1mm}
\resizebox{0.95\linewidth}{!}{%
\begin{tabular}{l|p{14.5cm}}
\toprule
\textbf{Task} 
& \multicolumn{1}{c}{\textbf{Definition and Formalism}} \\
\midrule

\textbf{FER} 
& \textbf{Facial Emotion Recognition:} Determines an emotion label using facial cues only. \\
& \quad \emph{Formal Definition:} \quad $x_{\mathrm{face}} \;\rightarrow\; E_{\mathrm{emotion}}$. \\

\midrule

\textbf{CAER} 
& \textbf{Context-Aware Emotion Recognition:} Identifies an emotion label by leveraging both facial and contextual cues. \\
& \quad \emph{Formal Definition:} \quad $[\,x_{\mathrm{face}},\,x_{\mathrm{context}}\,]\;\rightarrow\;E_{\mathrm{emotion}}$. \\

\midrule

\textbf{ER with LLMs} 
& \textbf{Emotion Recognition with Large Language Models:} Generates intermediate reasoning steps (e.g., chain-of-thought) before predicting the final emotion. \\
& \quad \emph{Formal Definition:} \quad $[\,x_{\mathrm{face}},\,x_{\mathrm{context}}\,]\;\rightarrow\;Z_{\mathrm{mediate}}^{1 \dots n}\;\rightarrow\;E_{\mathrm{emotion}}$. \\

\midrule

\textbf{HS} 
& \textbf{Humor Study:} Explains humor triggers in text or images, focusing on what makes a stimulus humorous. \\
& \quad \emph{Formal Definition:} \quad $H_{\mathrm{humor}}\;\rightarrow\;I_{\mathrm{humor}}$. \\

\midrule

\textbf{ECE} 
& \textbf{Emotion Cause Extraction:} Identifies specific triggers of a pre-given emotion from facial and contextual information. \\
& \quad \emph{Formal Definition:} \quad $[\,E_{\mathrm{emotion}},\,x_{\mathrm{face}},\,x_{\mathrm{context}}\,]\;\rightarrow\;T_{\mathrm{triggers}}$. \\

\midrule

\rowcolor{gray!20}
\textbf{EI} 
& \textbf{Emotion Interpretation:} Provides a broader and deeper explanation of an emotion’s triggers, potentially extending beyond observable cues. \\
& \quad \emph{Formal Definition:} \quad $[\,E_{\mathrm{emotion}},\,x_{\mathrm{face}},\,x_{\mathrm{context}}\,]\;\rightarrow\;I_{\mathrm{general\_trigger}}$. \\

\bottomrule
\end{tabular}
}
\vspace{-2mm}
\end{table*}

\subsection{Humor Study}
\label{subsec:humor_related_work}
Humor is a specialized affective phenomenon that has received extensive attention~\citep{cvpr_humor,hwang2023memecap,hessel2023androids,hyun2023smile,chinese_humor,urfunny,humorRecog,colbert,multimodal-humor}. These works investigate features eliciting laughter, from cartoon contexts~\citep{cvpr_humor} to internet memes~\citep{hwang2023memecap} and video laugh reasoning~\citep{hyun2023smile}. \citet{hessel2023androids} tested LLMs on a subset of the New Yorker Cartoon Caption Contest to see whether they grasp humor’s intricacies. While humor research constitutes a form of \emph{emotional interpretation}—aiming to elucidate what makes content funny—our approach is broader, targeting the triggers of various emotional states rather than focusing exclusively on amusement.

\subsection{Emotion Cause Extraction}
\label{subsec:ece_related_work}
\emph{Emotion Cause Extraction} (ECE) seeks to find textual or multimodal clues explaining a known emotion~\citep{ece,ecpe}. Early ECE work focused on identifying cause-effect pairs in textual corpora, often via multi-task learning to predict both emotion labels and their antecedents~\citep{ecpe}. Recently, \citet{wang2024semeval} extended ECE to a \emph{multimodal} setting in a SemEval challenge, where participants leveraged powerful LLM-based methods~\citep{zhang2024samsung,cheng2024mips,lei2023instructerc} to identify emotional triggers in speaker-centric conversations. Our \emph{Emotion Interpretation} framework is related to ECE but goes further: it does not simply locate a cause within the input; rather, it allows for generative, flexible triggers (including implicit or \emph{off-screen} context) and produces deeper explanations about \emph{why} an individual feels a specific emotion.

\subsection{Chain-of-Thought Prompting}
\label{subsec:cot_related_work}
Chain-of-thought (CoT) prompting improves problem-solving by prompting LLMs to articulate intermediate reasoning steps~\citep{self-ask,self-refine,tree-of-thought,graph-of-thoughts,mm-cot,auto-cot}. \citet{self-ask} introduced the \emph{Self-Ask} strategy, having LLMs generate and answer sub-questions. \citet{mm-cot} extended this approach to multimodal contexts by decoupling rationale generation and reasoning. Our \emph{Coarse-to-Fine Self-Ask} (CFSA) method similarly structures an LLM’s introspection but is specialized for \emph{emotion interpretation}, transitioning from general queries (e.g., number of people, basic context) to scenario-specific analysis of triggers. This hierarchical questioning strategy uncovers both explicit and implicit factors behind emotions, thus expanding CoT approaches into deeper affective reasoning.

%% file: sections/problem.tex
\section{Problem Definition}
\label{sec:definition}

\noindent
\textbf{Proposed Task.}
To explain \emph{why} a given emotion emerges, we introduce \emph{Emotion Interpretation} (EI). Let \(\mathcal{X}\) be the space of images, each image \(x \in \mathcal{X}\) consisting of a \emph{face} component \(x_{\text{face}}\) and a broader \emph{context} \(x_{\text{context}}\). Let \(\mathcal{E}\) be the set of possible emotions (e.g., \emph{happy}, \emph{unhappy}). We then define the \emph{query space}:
\begin{equation}
    \mathcal{Q} \;=\; \mathcal{X} \,\times\, \mathcal{E},
    \label{eq:queryspace}
\end{equation}
where each query \(q \in \mathcal{Q}\) is an ordered pair \((x,\,e)\). Rather than predicting \(e\), EI aims to generate a set of \emph{emotional triggers} \(T\). Let \(\mathcal{S}\) be the set of all possible triggers, encompassing both \emph{free-form textual explanations} (e.g., full sentences) and \emph{concise labels} (e.g., ``job loss''). Formally, we introduce a \emph{generative function}
\begin{equation}
    G: \mathcal{Q} \;\longrightarrow\; \mathcal{P}(\mathcal{S}),
    \label{eq:genfunc}
\end{equation}
where \(\mathcal{P}(\mathcal{S})\) denotes the power set of \(\mathcal{S}\). For a query \(q = (x,\,e)\in \mathcal{Q}\), the output
\begin{equation}
    T \;=\; G(x,\,e) \;\subseteq\; \mathcal{S}
    \label{eq:triggerset}
\end{equation}
represents the set of emotional triggers. Each trigger \(t_i \in T\) may be either a descriptive sentence (e.g., “He is sad because he lost his job.”) or a concise tag (e.g., “job loss”). If \(\mathcal{S} = \mathcal{S}_{\text{sent}} \,\cup\, \mathcal{S}_{\text{tags}}\), then \(t_i\in \mathcal{S}_{\text{sent}}\) (sentence-based explanations) or \(t_i \in \mathcal{S}_{\text{tags}}\) (concise labels). By letting \(T\subseteq \mathcal{S}\), we allow multiple triggers to coexist, thereby capturing a more nuanced explanation of an individual’s emotional state.

\noindent
\textbf{Emotional Triggers.}
We define an \emph{emotional trigger} as any stimulus \(\tau \in \mathcal{S}\) that elicits or modulates an individual’s emotional response. Typical examples of \(\tau\) include environmental elements \(\tau_{\text{env}}\) (e.g., a festive or tense atmosphere), social interactions \(\tau_{\text{social}}\) (e.g., conflicts, gatherings), physical cues \(\tau_{\text{phys}}\) (e.g., facial expressions, posture, gestures), and objects \(\tau_{\text{obj}}\) with sentimental value. While some triggers are directly observable, others emerge from less explicit or \emph{off-screen} factors (e.g., cultural norms or hidden backstories). Accounting for both \(\tau_{\text{explicit}}\) and \(\tau_{\text{implicit}}\) broadens EI’s ability to offer a richer, more holistic interpretation of emotional states.

\noindent
\textbf{Relation to Existing Tasks.}
In contrast to \(T_{\text{ER}}\) (i.e., \emph{Emotion Recognition}), which often uses facial or contextual inputs to classify an emotion label, \(T_{\text{EI}}\) (\emph{Emotion Interpretation}) explores \emph{why} a given emotion arises. This extends \(T_{\text{ECE}}\) (\emph{Emotion Cause Extraction}), which locates triggers for a known emotion \(E_{\mathrm{emotion}}\) but seldom permits flexible, generative explanations. Likewise, \(T_{\text{EMER}}\) (\emph{Explainable Multimodal Emotion Reasoning}) frequently depends on multi-class classification, limiting the variety of triggers it can represent. Lastly, \(T_{\text{HS}}\) (\emph{Humor Study})~\citep{hessel2023androids} is a specialized form of \(T_{\text{EI}}\) devoted to explaining comedic stimuli, underscoring the wider applicability of interpretative frameworks. Although modern \(T_{\text{ER}}\) methods may incorporate contextual information or Large Language Models (LLMs) with intermediate reasoning, they still focus on \emph{which} emotion is present rather than \emph{why} it emerges.

\noindent
\textbf{Illustrative Examples \& Comparisons.}
Table~\ref{tab:ec_example} demonstrates how EI interprets excitement and joy in an LGBT event. By parsing the user’s query and identifying pertinent triggers, the system explains \emph{why} the individual experiences a particular emotion, rather than merely detecting \emph{which} emotion is displayed. For a broader comparison against existing emotion-related tasks, Table~\ref{tab:related_work} details their respective objectives and input-output formalizations. Critically, EI focuses on causal triggers and reasons for emotional states, whereas most conventional approaches emphasize label prediction.

%% file: sections/benchmark.tex
\begin{table*}[t]
\centering
\footnotesize
\caption{\small 
Human evaluation results on \textbf{EIBench} annotation quality. Each cell shows \((\text{mean},\,\text{std},\,[\text{min},\,\text{max}])\). ``Overall'' denotes the aggregated rating across all emotion categories.
}
\label{tab:human_evaluation}
\vspace{-1mm}
\resizebox{0.85\linewidth}{!}{%
\begin{tabular}{lcccc c}
\toprule
\textbf{Satisfaction} 
& \textbf{Happy} 
& \textbf{Angry} 
& \textbf{Sadness} 
& \textbf{Excitement} 
& \textbf{Overall} \\
\midrule
\textbf{Person 1} 
& $(4.92, 0.27, [4,5])$ 
& $(4.90, 0.30, [4,5])$ 
& $(4.64, 0.83, [1,5])$ 
& $(4.98, 0.13, [4,5])$ 
& $(4.86, 0.46, [1,5])$ \\

\textbf{Person 2} 
& $(4.38, 0.62, [3,5])$ 
& $(4.62, 0.72, [2,5])$ 
& $(3.65, 1.31, [1,5])$ 
& $(4.58, 0.96, [1,5])$ 
& $(4.31, 0.94, [1,5])$ \\

\textbf{Person 3} 
& $(3.54, 0.63, [3,5])$ 
& $(4.08, 0.71, [2,5])$ 
& $(4.30, 0.75, [3,5])$ 
& $(4.39, 0.70, [2,5])$ 
& $(4.08, 0.70, [2,5])$ \\
\midrule
\textbf{Average} 
& $(4.28, 0.54, [3,5])$ 
& $(4.12, 0.98, [1,5])$ 
& $(4.61, 0.63, [2,5])$ 
& $(4.65, 0.69, [1,5])$ 
& $(4.42, 0.73, [1,5])$ \\
\bottomrule
\end{tabular}
}
\vspace{-3mm}
\end{table*}

\begin{table}[t]
\centering
\footnotesize
\caption{\small Fine-grained of emotions within each primary category.}
\label{tab:emo2fine_grained}
\vspace{-1mm}
\resizebox{\linewidth}{!}{%
\begin{tabular}{l l p{6.8cm}}
\toprule
\textbf{Category} & \textbf{Primary} & \textbf{Fine-Grained Emotions} \\
\midrule
\multirow{2}{*}{\textbf{Negative}}
& \textbf{Angry}
& Annoyed, agitated, upset, irritated, outraged, infuriated, hostile, concerned, frustrated, serious, displeased, mad, surprised, shocked, exhibit \\
\cmidrule(lr){2-3}
& \textbf{Sad}
& Forlorn, contemplative, unhappy, disheartened, dismal, solemn, sorrowful, somber, distress, miserable, discontent, upset, disappointment, distraught, displeased, frown, weary, frustration, loneliness, tragic, disappointed, melancholic, pain, injury \\
\midrule
\multirow{2}{*}{\textbf{Positive}}
& \textbf{Excite}
& Thrill, inspired, stimulate, incite, spur, smile, happy, raised, joyful, fascinating, enjoying, brightly, spark, enthusiasm, funny, intense, pleasant, feathery \\
\cmidrule(lr){2-3}
& \textbf{Happy}
& Smile, lighthearted, radiant, contented, pleased, spirited, cheerful, exhilarated, glad, blissful, energetic, joyful, optimistic, enjoying, positive, surprised \\
\bottomrule
\end{tabular}
}
\vspace{-2mm}
\end{table}

\begin{table}[t]
\centering
\footnotesize
\caption{\small 
Comparison of various emotion-related datasets. ER stands for \emph{Emotion Recognition}, EMER for \emph{Explainable Multimodal Emotion Recognition}, and EI for \emph{Emotion Interpretation}. ``Annotator'' indicates the number of individual annotators, ``Explainable'' denotes whether the dataset supports explanatory or causal annotations, and ``Has Complex Label'' refers to the presence of multi-layer or more nuanced labeling.
}
\label{tab:dataset_compare}
\vspace{-1mm}
\resizebox{\linewidth}{!}{%
\begin{tabular}{l c c c c c}
\toprule
\textbf{Dataset} 
& \textbf{Task} 
& \textbf{Annotator} 
& \textbf{Emotion Types} 
& \textbf{Explainable} 
& \textbf{Has Complex Label} \\
\midrule
CAER-S~\citep{caer-s} 
& ER 
& 6  
& 7  
& \xmark  
& \xmark  \\
DFEW~\citep{dfew} 
& ER 
& 3  
& 7  
& \xmark  
& \xmark  \\
RAF-DB~\citep{raf-db}
& ER 
& 315 
& 7  
& \xmark  
& \xmark  \\
HECO~\citep{heco}  
& ER 
& 13 
& 8  
& \xmark  
& \xmark  \\
EMOTIC~\citep{emotic} 
& ER 
& --  
& 26 
& \xmark  
& \xmark  \\
EmoSet~\citep{yang2023emoset} 
& ER 
& 10 
& 8  
& \cmark  
& \xmark  \\
MER2023 (EMER)~\citep{lian2023explainable} 
& EMER 
& 6  
& 7  
& \cmark  
& \xmark  \\
\rowcolor{gray!20}
EIBench 
& EI 
& 4  
& 4  
& \cmark  
& \cmark  \\
\bottomrule
\end{tabular}
}
\vspace{-3mm}
\end{table}

\begin{figure}[t]
    \centering
    \includegraphics[width=0.45\textwidth]{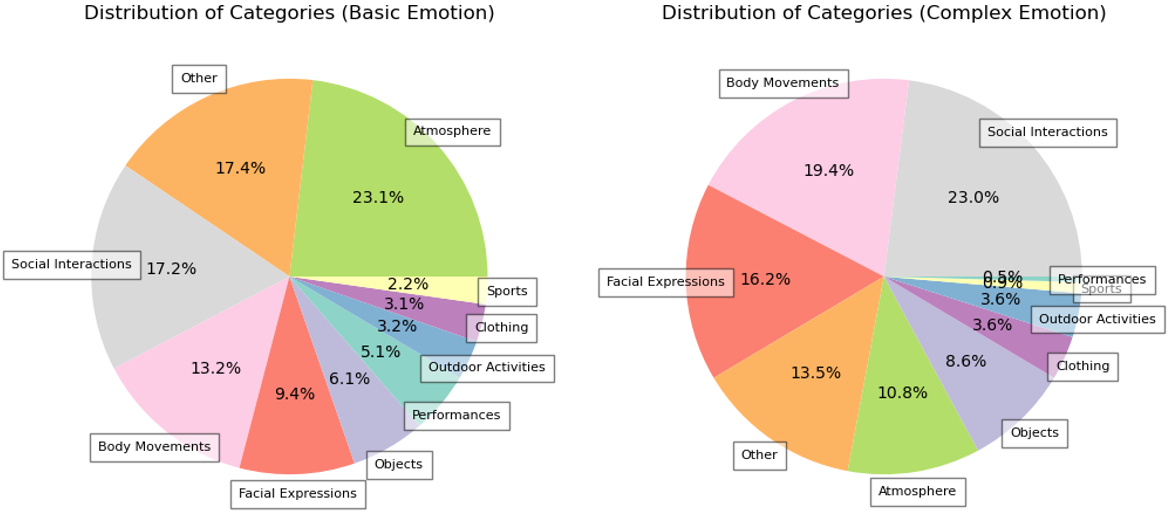}
    \caption{\small 
    Distribution of emotional triggers across distinct categories, contrasting \emph{Basic Emotions} (left) and \emph{Complex Emotions} (right). Each slice represents the proportion of triggers category.
    }
    \label{fig:frequency_triggers_basic_pie}
    \vspace{-0.10in}
\end{figure}

\section{Emotion Interpretation Benchmark}
\label{sec:ec_benchmark}
We now introduce \emph{EIBench}, a curated benchmark for EI that builds on CAER-S~\citep{caer-s} and EmoSet~\citep{yang2023emoset}. To the best of our knowledge, EIBench is the first dataset dedicated to explaining \emph{why} an emotion arises (rather than merely classifying \emph{which} emotion is present), featuring \num{1615} \emph{basic} EI samples and \num{50} \emph{complex} EI samples.

\subsection{VLLM-Assisted Dataset Construction}
\label{subsec:dataset_construction}
\noindent
\textbf{Coarse-to-Fine Annotation.}~As outlined in Appendix Figure~\ref{dataset_pipeline}, our \emph{Coarse-to-Fine Self-Ask} (CFSA) pipeline decomposes an initially implicit query into multiple simpler Visual Question Answering (VQA) tasks. CFSA involves four phases:
\emph{(1)~Initial Question Preprocessing},
\emph{(2)~General Self-Ask Thinking},
\emph{(3)~Scenario Self-Ask Thinking}, and
\emph{(4)~Emotion Summarization}.
After these automated steps, four volunteers thoroughly refine the annotations.

\noindent
\textbf{Initial Question Preprocessing.}~A concise prompt steers a Large Language Model (LLM), GPT-4 (denoted \(\phi\)), to enrich the user’s initial query \(s^{\mathrm{init}}\). Let \(s^{\mathrm{par}} = \phi(s^{\mathrm{init}})\). Given an image \(x_i \in \mathcal{X}\), we reconstruct a more elaborate prompt:
\[
  s_i^{\mathrm{rec}} \;=\; \texttt{llava}\bigl(x_i,\, s^{\mathrm{par}}\bigr),
\]
where \(\texttt{LLaVA-v1.6-34B}\) (\texttt{llava}) is a state-of-the-art Vision-Language Model. While such VLLMs capture many visual details, they tend to overlook subtle emotional cues~\citep{emotionllama}, necessitating the next ``self-ask'' phase.

\noindent
\textbf{General Self-Asking.}~We prompt GPT-4 to generate open-ended questions across the dataset, storing them in \(\mathcal{S}^{\mathrm{gen}}\). From \(\mathcal{S}^{\mathrm{gen}}\), we identify four frequently repeated queries, \(\mathcal{S}^{\mathrm{freq}}=\{s_1^{\mathrm{freq}}, s_2^{\mathrm{freq}}, s_3^{\mathrm{freq}}, s_4^{\mathrm{freq}}\}\), focusing on:
\emph{(i)~number of people},
\emph{(ii)~activities/interactions},
\emph{(iii)~facial expressions}, and
\emph{(iv)~body language}.
Each query \(s_j^{\mathrm{freq}}\) prompts \texttt{llava} to produce an answer \(a_j^{\mathrm{gen}}\), aggregated into \(\mathcal{A}^{\mathrm{gen}}\).

\noindent
\textbf{Scenario Self-Asking.}~We then supply the user query \(s^{\mathrm{query}}\), reconstructed prompt \(s_i^{\mathrm{rec}}\), and the pairs \(\{\mathcal{S}^{\mathrm{freq}}, \mathcal{A}^{\mathrm{gen}}\}\) to \texttt{llava} for scenario-level questioning, yielding \(\mathcal{S}_i^{\mathrm{sce}}\). Finally, an advanced LLM (e.g., LLaMA-3) integrates all collected answers to \emph{summarize} emotional triggers. Table~\ref{tab:ec_example} illustrates a CFSA-assisted annotation example.

\noindent
\textbf{Human In-the-Loop Annotation.}~The CFSA pipeline serves as a baseline. Four human annotators refine these automatic labels by:
\emph{(1)~removing hallucinations} (Appendix~\ref{hallucination}),
\emph{(2)~adding commonsense knowledge} (Appendix~\ref{commonsense}), and
\emph{(3)~pruning irrelevant triggers}.
To validate annotation quality, we randomly sample 50 images from each emotion category (200 total) for a final review by three volunteers, who rate their confidence in triggers on a 0--5 scale (scores < 3 signal poor or incomplete triggers). As shown in Table~\ref{tab:human_evaluation}, all final ratings exceed 4.0, demonstrating EIBench’s reliable annotations.

\subsection{Dataset Overview \& Evaluation}
\label{sec:evaluation_dataset}
\noindent
\textbf{Data Sources.}
EIBench is derived from CAER-S~\citep{caer-s}, which features seven emotion types (\emph{angry, disgust, fear, happy, neutral, sad, surprise}), and EmoSet~\citep{yang2023emoset}, comprising eight (\emph{anger, disgust, fear, sadness, amusement, awe, contentment, excitement}). To balance diversity with manageable annotation costs, we focus on four \emph{target} emotions: \emph{angry, sad, excited}, and \emph{happy}.

\noindent
\textbf{Data Composition \& Trigger Distribution.}
Table~\ref{tab:emo2fine_grained} lists fine-grained variants (e.g., \emph{annoyed}, \emph{forlorn}, \emph{thrilled}) for these four primary emotions. EIBench also includes 50 \emph{complex} samples, each annotated from multiple emotional perspectives. Emotional triggers fall into ten broad categories (e.g., \emph{atmosphere}, \emph{social interactions}, \emph{body movements}), as illustrated in Figure~\ref{fig:frequency_triggers_basic_pie}. Notably, \emph{atmosphere} and \emph{other} predominate for \emph{basic} emotions, while \emph{social interactions} and \emph{body movements} dominate the \emph{complex} subset.

\noindent
\textbf{Comparison with Existing Datasets.}
Table~\ref{tab:dataset_compare} contrasts EIBench with other emotion-related corpora. Unlike conventional datasets that classify a single dominant emotion, EIBench enables generative explanations of \emph{why} an emotion emerges, including \emph{complex} labeling. Appendix~\ref{sec:dataset_vis} provides further visualization of nuanced subset samples.

\noindent
\textbf{Evaluation Metrics.}~We measure performance via:
\emph{(1)~Emotional Trigger Recall}, which checks whether predicted triggers overlap with ground-truth annotations (multiple valid triggers can exist for one sample); and
\emph{(2)~Long-Term Coherence}, which assesses whether a model maintains thematic and emotional consistency in longer outputs. Specifically, we extract triggers from LLaMA-3 or ChatGPT3.5 responses, then use a BERT-based approach~\citep{devlin2018bert} to measure sentence-to-sentence similarity.

%% file: sections/experiment.tex
\begin{table*}[t]
\centering
\scriptsize
\caption{\small
Basic EI performance of open-source and closed-source language models on four emotion subclasses (\emph{Happy}, \emph{Angry}, \emph{Sadness}, \emph{Excitement}). 
Scores are reported under LLaMA-3 / ChatGPT criteria, with ``Overall'' denoting the aggregated result.
}
\label{tab:evaluation}
\vspace{-1mm}
\begin{tabular}{l|cccc|c}
\toprule
\textbf{Models} 
& \textbf{Happy} 
& \textbf{Angry} 
& \textbf{Sadness} 
& \textbf{Excitement} 
& \textbf{Overall} \\
\midrule

\multicolumn{6}{l}{\emph{User Question}} \\
\midrule

Qwen-VL-Chat    
& 32.09/39.68  
& 22.32/26.10  
& 30.64/33.88  
& 25.02/36.32  
& 26.45/33.65 \\

Video-LLaVA     
& 55.55/53.28  
& 40.42/36.97  
& 50.62/45.25  
& 51.78/52.23  
& 49.26/47.06 \\

MiniGPT-v2      
& 52.78/51.80  
& \textbf{47.10/47.76}  
& \textbf{60.47/58.14}  
& 50.78/53.66  
& 52.89/53.59 \\

Otter           
& 45.63/49.25  
& 42.53/43.07  
& 47.67/46.19  
& 39.47/48.30  
& 42.81/46.64 \\

LLaVA-1.5 (13B) 
& \textbf{59.01/57.52}  
& 45.44/41.88  
& 55.16/48.64  
& \textbf{57.46/58.73}  
& \textbf{54.37/52.20} \\

LLaVA-NEXT (7B)  
& 54.16/49.24  
& 43.71/39.87  
& 53.29/46.52  
& 58.90/53.06  
& 53.82/48.18 \\

LLaVA-NEXT (13B) 
& 57.17/55.18  
& 43.16/37.93  
& 54.16/45.42  
& 59.38/55.29  
& 54.33/48.79 \\

LLaVA-NEXT (34B) 
& 54.50/51.03  
& 38.96/35.65  
& 51.10/47.21  
& 51.77/52.04  
& 49.03/47.13 \\
\midrule

\multicolumn{6}{l}{\emph{User Question \& Caption}} \\
\midrule

Qwen-VL-Chat    
& 41.94/46.34  
& 32.71/31.91  
& 41.82/44.16  
& 38.65/43.84  
& 38.47/41.54 \\

Video-LLaVA     
& 56.77/58.79  
& 43.65/43.86  
& 54.25/55.12  
& 55.35/59.42  
& 52.63/54.85 \\

MiniGPT-v2      
& 55.11/60.04  
& 47.95/51.00  
& \textbf{62.29/64.24}  
& 51.55/57.90  
& 54.05/58.37 \\

Otter           
& 48.97/54.67  
& 34.22/37.12  
& 34.57/37.55  
& 35.27/42.99  
& 35.62/40.85 \\

LLaVA-1.5 (13B) 
& 57.91/58.46  
& 43.75/40.72  
& 55.47/51.46  
& 56.42/59.42  
& 53.55/53.13 \\

LLaVA-NEXT (7B)  
& \textbf{64.32/61.00}  
& 48.60/46.74  
& 58.75/53.00  
& \textbf{62.99/59.39}  
& 58.80/54.97 \\

LLaVA-NEXT (13B) 
& 61.99/61.95  
& \textbf{48.84/46.85}  
& 59.62/55.18  
& 62.17/59.95  
& \textbf{58.60/55.92} \\

LLaVA-NEXT (34B) 
& 57.51/62.73  
& 46.47/47.87  
& 58.35/55.84  
& 60.17/59.64  
& 56.60/56.24 \\

LLaMA-3 (8B) \scriptsize \textcolor{red}{(Text Only)}
& 52.36/50.73  
& 34.78/32.71  
& 52.29/46.87  
& 43.62/42.06  
& 44.73/41.94 \\
\midrule

\multicolumn{6}{l}{\emph{User Question \& CoT}} \\
\midrule

Qwen-VL-Chat    
& 41.99/44.46  
& 34.62/31.06  
& 43.64/39.30  
& 32.78/40.04  
& 36.79/38.18 \\

Video-LLaVA     
& 51.42/47.63  
& 42.68/35.65  
& 56.77/46.29  
& 53.01/46.98  
& 51.81/44.42 \\

MiniGPT-v2      
& 56.36/57.58  
& 47.71/48.32  
& \textbf{59.46/56.79}  
& 50.21/52.39  
& 52.67/53.08 \\

Otter           
& 49.97/51.91  
& 43.23/43.71  
& 50.15/46.86  
& 42.30/47.16  
& 45.17/46.61 \\

LLaVA-1.5 (13B) 
& \textbf{59.12/56.94}  
& 40.97/34.44  
& 53.07/45.66  
& 54.16/54.36  
& 51.34/47.80 \\

LLaVA-NEXT (7B)  
& 54.74/52.04  
& 44.61/41.93  
& 52.69/47.63  
& 52.78/47.60  
& 51.14/46.66 \\

LLaVA-NEXT (13B) 
& 50.91/50.35  
& 42.21/38.81  
& 54.66/49.42  
& 51.64/49.39  
& 50.47/47.21 \\

LLaVA-NEXT (34B) 
& 52.17/49.55  
& \textbf{48.35/44.45}  
& 55.97/50.55  
& \textbf{55.29/53.46}  
& \textbf{53.84/50.50} \\

\rowcolor{gray!30}
CFSA (LLaVA-NEXT (34B))
& 69.68/68.72  
& 61.08/61.14  
& 68.39/69.46  
& 72.63/70.31  
& 68.81/68.04 \\
\midrule

\multicolumn{6}{l}{\emph{Close-source Models}} \\
\midrule

Qwen-vl-plus\footnotemark[1]   
& 29.05/27.22  
& 23.58/17.89  
& 38.35/30.08  
& 30.09/26.87  
& 31.00/25.90 \\

ChatGPT-4V\footnotemark[2]   
& 52.30/55.74  
& 48.93/48.57  
& 45.00/44.42  
& 46.38/49.90  
& 46.86/48.58 \\

ChatGPT-4o\footnotemark[3]    
& 52.94/50.78  
& 42.12/35.33  
& 49.79/46.42  
& 53.48/54.53  
& 49.99/47.93 \\

Claude-3-haiku\footnotemark[4]
& \textbf{59.20/60.28}  
& \textbf{49.87/49.84}  
& \textbf{67.21/63.26}  
& \textbf{67.55/68.10}  
& \textbf{63.24/62.41} \\

Claude-3-sonnet\footnotemark[4]
& 44.58/44.45  
& 38.95/42.86  
& 55.98/54.40  
& 61.41/62.24  
& 54.10/54.89 \\

\bottomrule
\end{tabular}
\vspace{-3mm}
\end{table*}

\footnotetext[1]{\url{https://qwenlm.github.io/blog/qwen-vl/}}
\footnotetext[2]{\url{https://openai.com/index/gpt-4-research/}}
\footnotetext[3]{\url{https://openai.com/index/hello-gpt-4o/}}
\footnotetext[4]{\url{https://docs.anthropic.com/en/docs/models-overview}}

\begin{table}[t]
\centering
\footnotesize
\caption{\small
Effect of persona prompts on model performance, evaluated by LLaMA-3 / ChatGPT criteria. 
``\textbf{w/o Persona}'' indicates no explicit persona, while 
``\textbf{AI Assistant}, \textbf{Architecture}, \textbf{Emotion}'' specify distinct persona setups.
}
\label{tab:persona}
\vspace{-1mm}
\resizebox{\linewidth}{!}{%
\begin{tabular}{lcccc}
\toprule
\textbf{Model} 
& \textbf{w/o Persona}
& \textbf{AI Assistant}
& \textbf{Architecture}
& \textbf{Emotion} \\
\midrule
LLaVA-NEXT (7B) 
& 52.09/46.64 
& 49.48/46.13 
& 45.32/38.40 
& \textbf{53.82/48.18} \\

LLaVA-NEXT (13B) 
& 52.44/50.07 
& 49.69/48.12 
& 44.26/35.79 
& \textbf{54.33/48.79} \\

LLaVA-1.5 (13B) 
& 51.58/53.62 
& 51.04/50.66 
& 49.58/43.16 
& \textbf{54.37/52.20} \\

Claude-3-haiku 
& 58.28/58.62 
& 60.37/59.86 
& 31.81/25.53 
& \textbf{63.24/62.41} \\
\bottomrule
\end{tabular}
}
\vspace{-3mm}
\end{table}

\begin{table*}[t]
\centering
\begin{minipage}[t]{0.53\textwidth}
\centering
\vspace{-4mm}
\caption{\small
Evaluation of complex EI ability across various VLLMs. 
Scores denote \emph{Recall} under LLaMA-3 / ChatGPT criteria.
}
\label{tab:complex_emo}
\vspace{1mm}
\scalebox{0.85}{%
\begin{tabular}{l c}
\toprule
\textbf{Models} & \textbf{Recall} \\
\midrule
\multicolumn{2}{l}{\emph{Open-Source}} \\
Qwen-VL-Chat      & 22.00/32.40 \\
Video-LLaVA       & 30.90/32.27 \\
MiniGPT-v2        & 35.10/36.00 \\
Otter             & 27.90/33.23 \\
LLaVA-1.5 (13B)   & \underline{38.10/39.53} \\
LLaVA-NEXT (7B)   & 38.71/33.50 \\
LLaVA-NEXT (13B)  & 39.16/33.60 \\
LLaVA-NEXT (34B)  & 35.37/33.10 \\
\midrule
\multicolumn{2}{l}{\emph{Close-Source}} \\
Qwen-vl-plus      & 20.37/19.60 \\
Claude-3-haiku    & 24.00/24.77 \\
Claude-3-sonnet   & 21.37/22.40 \\
ChatGPT-4V        & 28.00/30.60 \\
ChatGPT-4o        & \textbf{39.27/39.57} \\
\bottomrule
\end{tabular}
}
\end{minipage}
\hfill
\begin{minipage}[t]{0.45\textwidth}
\centering
\vspace{-4mm}
\caption{\small 
Long-Term Coherence among VLLMs for the \emph{User Question} setting.
Values are BERT-based similarity scores.
}
\label{tab:Long_term_Coherence}
\vspace{1mm}
\scalebox{0.85}{%
\begin{tabular}{l c}
\toprule
\textbf{Models} & \textbf{Coherence} \\
\midrule
\multicolumn{2}{l}{\emph{Open-Source}} \\
Qwen-VL-Chat     & 84.49 \\
Video-LLaVA      & 84.89 \\
MiniGPT-v2       & 84.70 \\
Otter            & \underline{85.03} \\
LLaVA-1.5 (13B)  & 84.50 \\
LLaVA-NEXT (7B)  & 81.02 \\
LLaVA-NEXT (13B) & 81.09 \\
LLaVA-NEXT (34B) & 84.96 \\
\midrule
\multicolumn{2}{l}{\emph{Close-Source}} \\
Qwen-vl-plus     & 83.00 \\
Claude-3-haiku   & \textbf{85.98} \\
Claude-3-sonnet  & 84.53 \\
ChatGPT-4V       & 81.97 \\
ChatGPT-4o       & 80.65 \\
\bottomrule
\end{tabular}
}
\end{minipage}
\vspace{-4mm}
\end{table*}

\section{Experiments}
\label{sec:experiments}
In this section, we evaluate both prominent open-source and proprietary models on our proposed benchmark.~We design four distinct modes to assess each model’s capability in \emph{Emotion Interpretation} (EI), and we conclude with an in-depth analysis of these results.

\subsection{Experimental Setup}
\label{subsec:experimental_setup}

\noindent
\textbf{Modes of Evaluation.}
We introduce four modes to investigate how LLMs approach EI:
\begin{itemize}
     \item \emph{User Question (UQ)}:
    In this zero-shot scenario, the user’s question is provided verbatim. This setting examines each model’s direct ability to handle natural, potentially ambiguous queries.
    
    \item \emph{User Question + Caption (UQ+C)}:
    The user question is enriched by a caption (see Section~\ref{sec:ec_benchmark} for details on caption generation). This aims to clarify context and improve accuracy. We also include a text-only baseline with LLaMA-3 fed the same caption.
    
    \item \emph{User Question + CoT (UQ+CoT)}:
    In this mode, a succinct chain-of-thought style prompt (e.g., “Let’s think step by step”) is appended to the user’s query. This setup intentionally encourages the model to reason more systematically, revealing key intermediate thought processes.

    \item \emph{CFSA Setting (CFSA)}:
    We carefully employ the Coarse-to-Fine Self-Ask (CFSA) method, implemented by LLaVA-NEXT (34B), to divide the EI task into more manageable sub-queries.~This scenario essentially demonstrates an upper-bound performance facilitated by a well-structured question–answer pipeline.
\end{itemize}

\subsection{Overall Performance}
\label{subsec:overall_performance}

\noindent
\textbf{Basic EI Results.}
Table~\ref{tab:evaluation} presents the scores of various models—open-source and closed-source—on the four primary emotion categories (\emph{Happy}, \emph{Angry}, \emph{Sadness}, \emph{Excitement}). Among open-source models, the LLaVA family and MiniGPT-v2 generally excel, with Qwen-VL-Chat consistently lagging. Notably, \emph{Video-LLaVA} and \emph{Otter} occupy mid-tier performance, although Otter underperforms significantly in \emph{Excitement}. Closed-source systems, particularly the \emph{Claude-3} series and \emph{ChatGPT-4o}, typically surpass open-source approaches in the direct user-question setting. The Qwen-vl-plus, however, performs poorly compared to other closed-source alternatives.

\noindent
\textbf{Complex EI Results.}
Moving to complex EI, Table~\ref{tab:complex_emo} shows model recall on our multifaceted subset. Here, top open-source models (e.g., \emph{LLaVA-1.5} at 38.10/39.53) come close to \emph{ChatGPT-4o}, the best closed-source model in these scenarios. Interestingly, although \emph{Claude-3} variants dominate simpler EI tasks, they do not achieve top-tier results on these more complex samples. This discrepancy suggests that while Claude-3 excels at single-perspective (basic) EI, it struggles with the additional demands of deeper multi-perspective emotional contexts.

\noindent
\textbf{Long-Term Coherence.}
Table~\ref{tab:Long_term_Coherence} evaluates \emph{long-term coherence} via a BERT-based similarity measure. Most models cluster around an 80--86\% range, demonstrating the capacity to maintain thematic or emotional consistency across longer outputs. Although coherence scores are relatively high overall, they do not necessarily translate into superior EI performance—underscoring that logical textual flow can be partially decoupled from accurate emotional insight.

\subsection{Ablation on Persona Prompts}
\label{subsec:persona_ablation}
Inspired by PsychoBench~\citep{persona_prompt}, we examine whether assigning different \emph{personas} to LLMs modulates EI performance. Table~\ref{tab:persona} compares four settings: 
\emph{(i) no persona},
\emph{(ii) AI Assistant persona}, 
\emph{(iii) Architecture expert}, and
\emph{(iv) Emotion expert}. 
Models consistently achieve higher scores when framed as \emph{emotion} experts, suggesting domain-specific personas help center chain-of-thought on emotional triggers. In contrast, an \emph{architecture} persona often degrades EI performance below the no-persona baseline, implying mismatched prompts overshadow emotional reasoning. These results show that well-chosen personas, aligned with the target domain, can guide LLMs toward more accurate, context-driven EI interpretations.

\subsection{Analysis of Evaluation Modes}
\label{subsec:modes_analysis}

\noindent
\textbf{User Question vs. Caption.}
Comparing \emph{UQ} to \emph{UQ+C} (Table~\ref{tab:evaluation}) reveals that providing a relevant caption consistently boosts performance, often by 2--5 points. The exception is \emph{Otter}, whose score drops when a caption is added, possibly because the additional text conflicts with its internal reasoning framework. Meanwhile, \emph{MiniGPT-v2} gains substantially in \emph{Angry} and \emph{Sadness}, whereas the \emph{LLaVA} variants post the highest overall figures. Interestingly, scaling the LLaVA models to 34B does not yield a clear advantage—both 7B and 13B configurations can achieve competitive or better results in certain subsets.

\noindent
\textbf{Chain of Thought Prompting.}
Adopting \emph{UQ+CoT} (cf. Table~\ref{tab:evaluation}) generally improves performance over \emph{UQ}, indicating that a structured, step-by-step approach helps surface hidden emotional triggers. These gains align with the CFSA pipeline’s rationale that detailed introspection (i.e., CoT) better exposes causal factors behind human emotions. Indeed, the higher performance in CoT-like settings further supports the idea that complex tasks—like explaining \emph{why} a person feels a certain way—benefit more from reasoned dialogues than from direct, single-shot responses.

\noindent
\textbf{CFSA Upper Bound.}
By converting queries into multiple simpler VQA tasks, the \emph{CFSA} configuration yields the strongest results among open-source VLLMs, capturing around 68\% of emotional triggers in Table~\ref{tab:evaluation}. This still falls short of manual annotations, highlighting the complexity of EI. Nonetheless, it demonstrates that \emph{a carefully structured pipeline} can significantly narrow the gap between raw zero-shot performance and a more expert-level approach.

\subsection{Key Observations and Limitations}
\label{subsec:observations_limitations}

\noindent
\textbf{Human-Level Annotation Gap.}~While CFSA-based methods show promise, they still exhibit a noticeable gap from human-labeled data, indicating that subtle emotional cues remain difficult for LLMs to capture. This gap reinforces the need for refined instruction tuning and more sophisticated context modeling.

\noindent
\textbf{Discrepancies Across Emotions.}~Both Table~\ref{tab:evaluation} and Table~\ref{tab:complex_emo} reveal that performance varies widely by emotion category. Models generally handle \emph{Happy} and \emph{Sadness} more successfully, whereas \emph{Excitement} and \emph{Complex Mixed Emotions} pose greater challenges—possibly due to more nuanced or overlapping triggers.

\noindent
\textbf{Open vs.\ Closed-Source Trade-offs.}~Although certain open-source systems (e.g., LLaVA-1.5, LLaVA-NEXT) rival or surpass smaller closed-source models, they still typically trail behind top-tier closed-source ones (e.g., Claude-3, ChatGPT-4o). This discrepancy emphasizes how additional proprietary data and advanced training can drive incremental performance gains.

%% file: sections/conclusion.tex
\section{Conclusion}
\label{sec:discussion}
This work reframes emotion analysis by asking \emph{why} an emotion arises rather than \emph{which} emotion is present. We introduced EIBench for \emph{Emotion Interpretation (EI)}, highlighting causal triggers of affective states via both explicit cues (e.g., visible objects) and implicit factors (e.g., cultural norms). Our Coarse-to-Fine Self-Ask pipeline and evaluations on open-source and proprietary large language models demonstrate the potential of EI to enrich empathy and context-awareness in AI. Nevertheless, models still struggle with overlapping emotions and subtle cues beyond their training scope, our dataset, though broad, cannot capture all real-world scenarios, and existing interpretability metrics for causal reasoning need further refinement. Future work should explore deeper integration with audio and textual dialogues, extended causal modeling to handle subtle emotional overlaps, and more adaptive evaluation protocols in dynamic contexts \emph{with user-specific adaptability}.

%% file: Supplementary/dataset_and_code_access.tex
\section{EIBench}
\label{supp:eibench}

\subsection{Practical Applications}
\label{supp:application}
EIBench’s emphasis on \emph{Emotion Interpretation (EI)} supports a variety of real-world use cases:

\begin{enumerate}[leftmargin=12pt]
    \item \textbf{Enhanced Emotion Recognition:}
    Most datasets label emotions but ignore \emph{why} they occur. EIBench illuminates causal factors, further refining both accuracy and empathy in emotion recognition. Possible applications: customer service bots, mental health diagnostics, and interactive media, where \emph{causal} triggers foster more context-aware responses.
    
    \item \textbf{Adaptive Human-Computer Interaction (HCI):}
    Capturing \emph{why} users feel certain emotions, EIBench-trained models provide adaptive, personalized experiences. Virtual assistants, interactive gaming, or user-facing platforms can tailor responses to precise emotional contexts.
    
    \item \textbf{Psychological and Behavioral Studies:}
    Researchers can use EIBench’s triggers to uncover patterns in emotional responses and factors shaping them. These insights inform clinical psychology interventions and broaden our grasp of human behavior.

    \item \textbf{Deeper Social Media Analysis:}
    EIBench extends sentiment analysis by unveiling the emotional context behind online posts. This expanded layer of interpretation aids brands and organizations in tracking public sentiment more accurately, responding to feedback effectively, and managing their online presence with greater nuance.
\end{enumerate}

\subsection{Intended Audiences}
EIBench aims to advance EI by capturing the subjective nature of emotional states. Addressing the dataset’s challenges can lead to \emph{empathetic} AI systems, enriching emotion-driven applications and enhancing human–computer interactions. Additionally, these insights may benefit tasks like humor understanding, harmful stance detection, and other domains that hinge on implicit emotion cues. Overall, EIBench paves the way for multifaceted, context-driven emotion interpretation, pushing the boundaries of next-generation EI research.

%% file: Supplementary/baseline_models.tex
\section{Baseline Models}

\subsection{Open-Source Models}
\label{sec:open_source_models}

\noindent \textbf{Qwen-VL-Chat.}~Qwen-VL-Chat~\cite{qwen} is a multimodal large language model (LLM)-based assistant developed by Alibaba Cloud. It manages multiple image inputs, multi-round question answering, and uses bounding boxes for grounding. Through a 448$\times$448-resolution visual encoder, Qwen-VL-Chat supports finer text recognition, document QA, and bounding box annotation. Additionally, it operates in English, Chinese, and other languages, enabling end-to-end recognition of bilingual text. Multi-image interleaved conversations allow image-to-image comparisons, enabling scenario analysis and multi-image storytelling.

\noindent \textbf{Video-LLaVA.}~Video-LLaVA~\cite{vllava} acts as a baseline for Large Vision-Language Models (LVLMs) that handle both images and videos within a unified visual feature space. By aligning image and video representations, Video-LLaVA allows models to enhance performance across both modalities simultaneously, often outperforming methods restricted to either static images or video alone.

\noindent \textbf{MiniGPT-v2.}~MiniGPT-v2~\cite{minigpt4v2} is a versatile multimodal model supporting diverse vision-language tasks such as image description, VQA, and grounding. It reduces visual token sequence length by merging adjacent tokens, thus enhancing training efficiency at high resolutions. Trained in three stages—broad pretraining, task-specific fine-tuning on high-quality datasets, and multimodal instruction tuning—MiniGPT-v2 excels at chatbot-style interactions and complex multimodal tasks.

\noindent \textbf{Otter.}~Otter~\cite{li2023otter} leverages \emph{OpenFlamingo}~\cite{openflamingo} to perform multi-modal in-context instruction tuning. Each data instance in its \emph{MIMIC-IT}~\cite{mimic} training set comprises an instruction-image-answer triplet along with relevant in-context examples. By conditioning the language model on image-caption or instruction-response pairs, Otter attains strong instruction-following skills and effectively learns from contextual exemplars.

\noindent \textbf{LLaVA-1.5.}~LLaVA-1.5~\cite{llava} builds on CLIP-ViT-L-336px~\cite{radford2021learning} with an additional MLP projection layer and integrates academic-task-focused VQA data. Compared to the original LLaVA, this version enhances cross-modal connections via an MLP connector and utilizes a broader set of VQA data. The 13B checkpoint for LLaVA-1.5 relies on around 1.2M publicly available data samples.

\noindent \textbf{LLaVA-NEXT.}~Relative to LLaVA-1.5, LLaVA-NEXT~\cite{liu2024llava} improves reasoning, optical character recognition (OCR), and world knowledge under high-resolution settings, reducing model hallucinations and capturing intricate image details. Training includes High-quality User Instruct Data and Multimodal Document/Chart Data, plus the flexibility to employ various LLM backbones (e.g., Mistral-7B~\cite{mistral} or Nous-Hermes-2-Yi-34B\footnotemark[1]).

\footnotetext[1]{\url{https://huggingface.co/NousResearch/Nous-Hermes-2-Yi-34B}}

\subsection{Close-Source Models}
\label{sec:closed_source_models}

\noindent \textbf{Qwen-vl-plus.}~Qwen-vl-plus expands on Qwen-VL’s capabilities for detailed recognition, text detection, and high-resolution image handling (e.g., millions of pixels, arbitrary aspect ratios). It performs competitively on a broad spectrum of visual tasks but is available only via an online API.

\noindent \textbf{Claude-3.}~Claude-3 from Anthropic underscores safety, controllability, and ethics—distinguishing it from ChatGPT via adversarial training that reduces bias and harmful outputs. Although ChatGPT also addresses safety, Claude emphasizes robust security measures and transparent documentation. While ChatGPT excels at broad NLP tasks, Claude’s stringent ethical guidelines may favor use cases requiring higher compliance standards.

\noindent \textbf{ChatGPT-4.}~ChatGPT-4 (ChatGPT-4o, ChatGPT-4V) is OpenAI’s state-of-the-art LLM, proficient in text generation, conversation, translation, summarization, and more. It incorporates extensive pretraining to boost coherence and fluency.~Like Claude, ChatGPT-4 has significant safety mechanisms for mitigating bias and harm, plus user-feedback loops to enhance performance. Its adaptability makes it effective for a wide array of applications, balancing general NLP strength with ethical safeguards.

\subsection{Basic EIBench}
\label{supp:basic_eibench}
EIBench is composed of two primary subsets—\emph{Basic} and \emph{Complex}. The \emph{Basic} subset contains \num{1615} samples, each aligned with one of four primary emotion categories (\emph{angry}, \emph{sad}, \emph{happy}, \emph{excited}). Unlike the \emph{Complex} subset, which may feature overlapping or multilayered emotions, the \emph{Basic} portion focuses on a single dominant emotion per instance. This design choice allows models to learn and generalize from relatively direct emotional triggers before grappling with more intricate scenarios.

\noindent \textbf{Annotation Approach.}~We follow the same \emph{Coarse-to-Fine Self-Ask (CFSA)} pipeline as outlined in the main text. However, unlike \emph{Complex} scenarios—where multiple viewpoints or confounding cues might need iterative clarification—the \emph{Basic} subset typically converges on a single, primary trigger. Consequently, annotators can identify and refine emotional cues (e.g., facial expressions, objects, or contextual details) in fewer self-ask rounds, thus ensuring the reliability of each final annotation.

\noindent \textbf{Scope and Limitations.}~Although each \emph{Basic} sample focuses on one principal emotion, subtler undertones (e.g., mild frustration coexisting with sadness) can still arise. Annotators are instructed to emphasize the dominant emotion, but residual emotional nuances may remain. Models trained on the \emph{Basic} subset alone often handle straightforward triggers well (e.g., “waiting in a queue,” “a celebratory event”), yet may perform less effectively when encountering real-world complexities or mixed emotional contexts—challenges that are central to \emph{Complex} EIBench.

\noindent \textbf{Intended Use.}~The \emph{Basic} subset is especially suited for initial baseline training, providing a gentle introduction for models to learn one dominant emotional cue per instance. Researchers can compare baseline performances on simpler triggers with the more layered triggers in the \emph{Complex} subset. Additionally, the straightforward, readily identifiable causes in the \emph{Basic} portion benefit educational demonstrations, helping novices grasp core mechanisms of emotion interpretation before tackling more advanced material. Overall, \emph{Basic EIBench} offers a structured entry point to explain \emph{why} a single emotion dominates a scene, complementing EIBench’s broader aim of preparing models for more nuanced, overlapping emotional states.

\subsection{Complex EI Subset}
\label{sec:dataset_vis}
In contrast to the \emph{Basic} subset, the \emph{Complex} EI subset comprises \num{50} samples featuring overlapping or multilayered emotions (e.g., joy mixed with regret, anger intertwined with concern). Such scenarios push models to identify multiple coexisting triggers and navigate nuanced social or cultural cues (Figure~\ref{trigger_shows}(e)).

\noindent \textbf{Scope and Design.}
Each \emph{Complex} instance often involves layered triggers (e.g., work-related stress combined with family conflict), requiring multi-step reasoning; interwoven perspectives (e.g., two individuals each experiencing distinct emotional reactions), which force the model to untangle different motivations; and implicit contextual depth (e.g., cultural practices or off-screen backstories) that may not appear explicitly but remain crucial for understanding the emotional state.

\noindent \textbf{Annotation Method.}
Compared to \emph{Basic} cases, annotators adopted a more iterative \emph{Coarse-to-Fine Self-Ask} flow to clarify overlapping cues and verify multiple triggers. This extra step ensures the final annotations encompass all relevant factors (e.g., social tension plus personal grief), rather than focusing on just the first visible cause.

\noindent \textbf{Impact and Utility.}
The \emph{Complex} subset highlights realistic emotional intricacies, fostering development of more robust \emph{Emotion Interpretation (EI)} models. Beyond academic interest, these examples aid use cases in mental health diagnostics and advanced HCI, where single-label assumptions fail to capture genuine emotional complexity. Together with the \emph{Basic} subset, these intricate scenarios enable a broader transition from straightforward emotion labeling to richer, more nuanced emotional understanding.

\section{Human-in-the-Loop Data Cleaning}
\label{sec:human_loop_cleaning}

\subsection{Addressing Hallucinations in VLLMs}
\label{hallucination}
Vision Large Language Models (VLLMs) can sometimes produce \emph{hallucinated} triggers unrelated to the actual image content. Table~\ref{tab:hullucination} shows examples in which the model invents triggers (e.g., “Doing mountain biking”) with no supporting evidence. Such hallucinations undermine dataset quality by misrepresenting the visual context. To mitigate these errors, we implement a human-in-the-loop cleaning process: annotators review the VLLM’s outputs, remove triggers not clearly supported by the image, and note ambiguous regions for further inspection. By systematically weeding out these misinterpretations, we reduce biases introduced by VLLM-driven hallucinations.

\subsection{Incorporating Commonsense Knowledge}
\label{commonsense}
Even when models avoid overt hallucinations, they may overlook \emph{commonsense} cues essential to explaining an emotional state. Table~\ref{tab:common_sense} illustrates how human annotators augment triggers with contextual or cultural knowledge absent from raw VLLM outputs. For instance, the model may label an emotion as “angry” but omit a crucial real-life cause (e.g., “waiting for lost luggage”), prompting annotators to add relevant details. By explicitly integrating commonsense reasoning, the final dataset more closely aligns with real-world emotional triggers, thus enhancing the fidelity and utility of EIBench for emotion interpretation tasks.

\section{Case Study of the VLLMs’ EI Abilities}
\label{case_study_of_VLLMs}
In this section, we present a detailed examination of how various Vision-Language Models (VLLMs) handle \emph{Emotion Interpretation (EI)}, focusing on both \emph{hallucinations} and \emph{commonsense knowledge integration}. Tables~\ref{tab:hullucination} and~\ref{tab:common_sense} illustrate how a human-in-the-loop data cleaning process identifies and corrects inaccuracies or omissions in VLLM outputs.

\noindent
\textbf{Hallucinations in VLLMs.}
Table~\ref{tab:hullucination} shows instances where the VLLM-generated triggers deviate from the image content (e.g., ``\emph{Doing mountain biking}'' when no bike is present), misrepresenting the scene and undermining dataset quality. By having human annotators remove or adjust these erroneous details, we mitigate biases that might otherwise skew emotion interpretation.

\noindent
\textbf{Commonsense Knowledge Integration.}
Table~\ref{tab:common_sense} highlights cases where VLLMs lack crucial background context (e.g., ``\emph{first Halloween experience},'' ``\emph{first time to Beijing}''). Human annotators augment these triggers with necessary cultural or situational information, yielding more realistic and representative data annotations.

\begin{figure*}[h]
\centering
\includegraphics[width=0.9\textwidth]{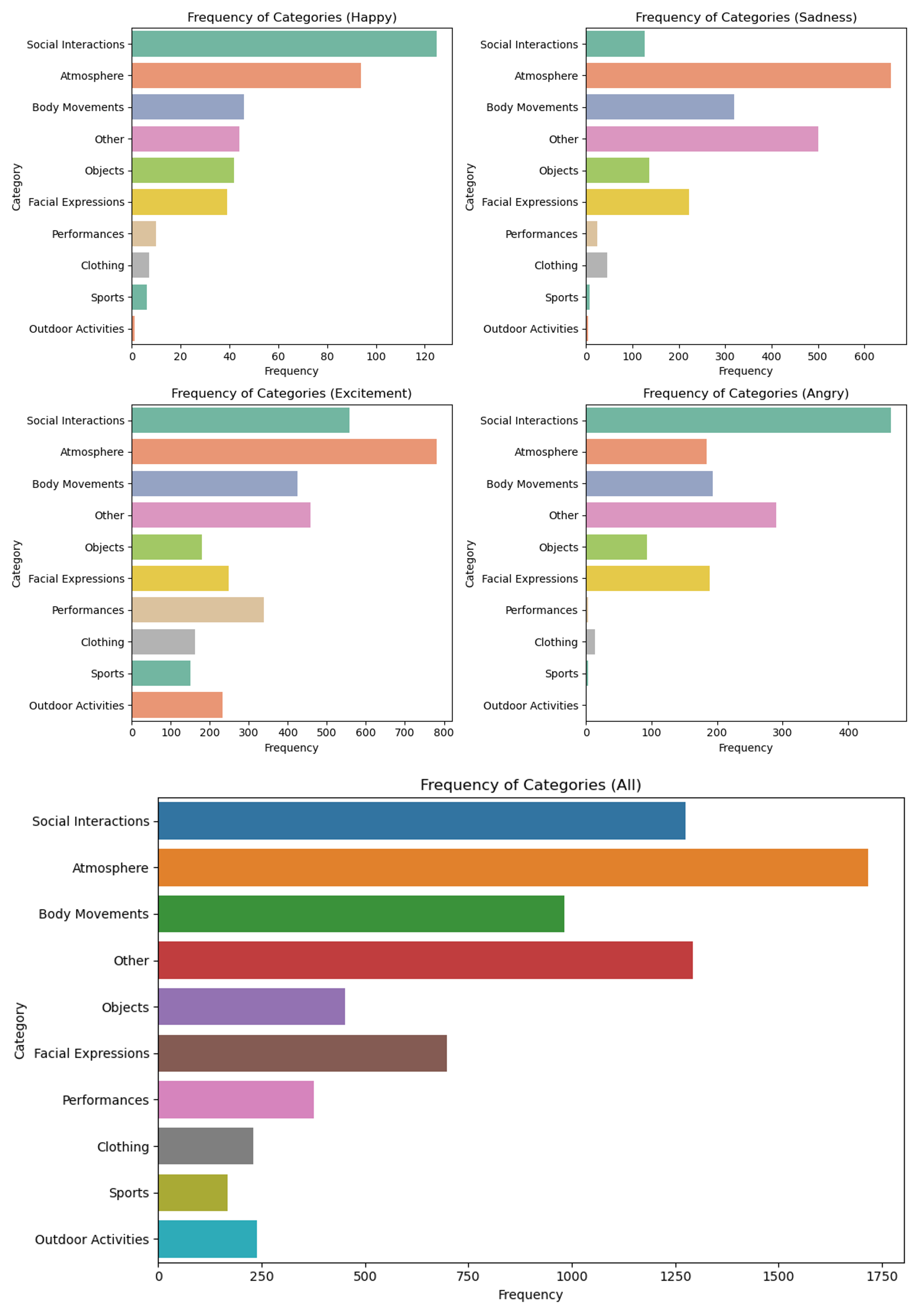}
\vspace{-4mm}
\caption{\small Visualization of the numbers of emotional triggers across different categories (Basic Emotions).}
\label{frequency_triggers}
\vspace{-3mm}
\end{figure*}

\begin{figure*}[h]
\centering
\includegraphics[width=0.9\textwidth]{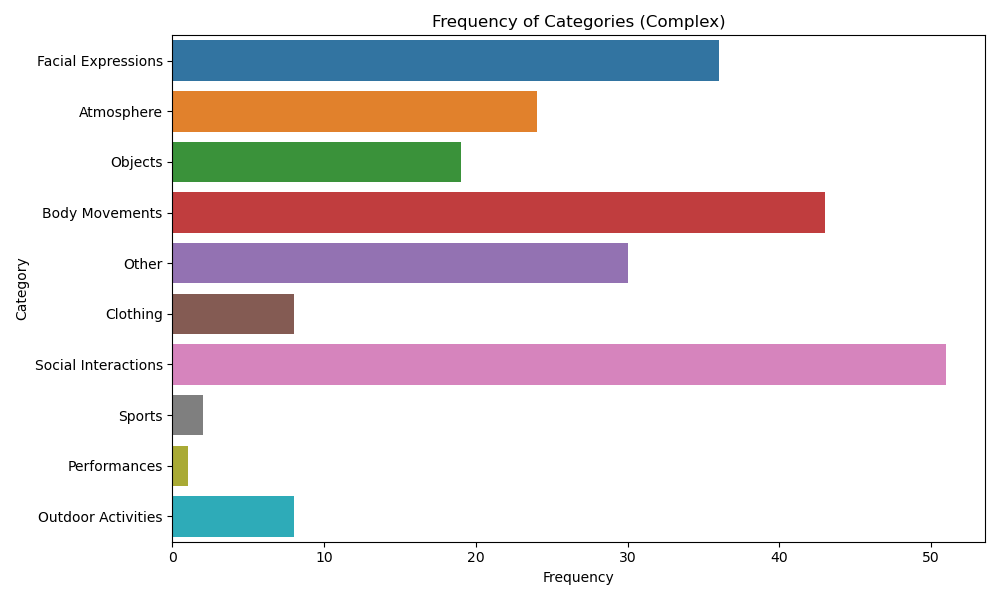}
\vspace{-4mm}
\caption{\small Visualization of the numbers of emotional triggers in the Complex EI Subset.}
\label{frequency_triggers_complex}
\vspace{-3mm}
\end{figure*}

\noindent
\textbf{Basic vs.\ Complex EI.}
Figures~\ref{frequency_triggers} and~\ref{frequency_triggers_complex} and the accompanying tables illustrate how emotional triggers distribute across \emph{Basic} and \emph{Complex} subsets. In simpler, single-emotion scenarios (Table~\ref{tab:app_basic_dataset_vis}), VLLMs often identify straightforward triggers (e.g., ``\emph{long wait},'' ``\emph{enjoying the view}''). Meanwhile, \emph{Complex} samples (Table~\ref{tab:app_com_dataset_vis}) feature overlapping triggers or multiple emotional states, frequently exposing model challenges in capturing less obvious cues.

\noindent
\textbf{Detailed Model Responses.}
Tables~\ref{tab:hullucination}--\ref{tab:common_sense} present user queries and ground-truth triggers, alongside raw VLLM outputs (e.g., Qwen-VL-Chat, LLaVA family, MiniGPT, Otter, and ChatGPT-4). Each response is evaluated by LLaMA-3 and ChatGPT for alignment with the annotated triggers. A common pattern emerges: 
Certain triggers (\emph{e.g.}, metal claws, intense gaze) are detected reliably, while subtler elements (\emph{e.g.}, wide-opening eyes, “defending gesture,” “shrunk muscle”) are overlooked or inconsistently recognized. Some VLLMs also invent erroneous triggers (e.g., “\emph{concern about a meal he’s preparing}”) incongruent with the annotated details.

\noindent
\textbf{Insights and Implications.}
These case studies highlight the complexity of moving from mere emotion \emph{recognition} to \emph{interpretation}. Straightforward triggers are typically recognized, but nuanced emotions often hinge on contextual, cultural, or implicit cues. Human review and data cleaning (Sections~\ref{hallucination}--\ref{commonsense}) remain vital for honing outputs, particularly in ambiguous or subtle contexts. EIBench thus provides a structured environment for testing not only \emph{Basic} scenarios but also the \emph{Complex} interactions that more closely mirror real-world emotional landscapes.

\begin{figure*}[h]
\centering
\includegraphics[width=1\textwidth]{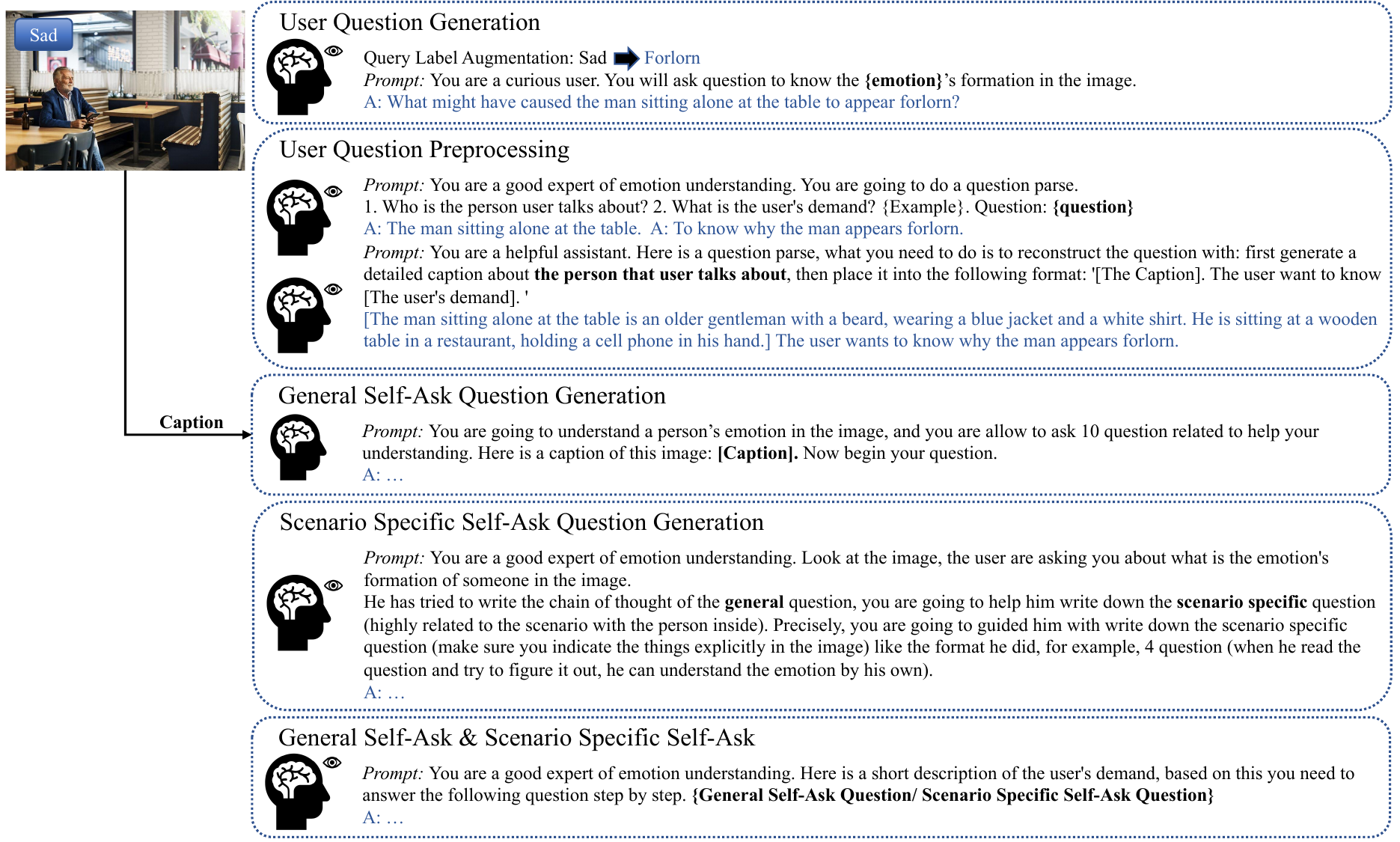}
\caption{\small Pipeline of the VLLM-assisted dataset construction.}
\label{dataset_pipeline}
\vspace{-4mm}
\end{figure*}

\begin{table*}[h]
\begin{minipage}{1\textwidth}
\centering
\vspace{-2mm}
\caption{\small Visualization of basic EI dataset, an image is corresponded to one user questions.}
\scalebox{0.65}{
    \begin{tabular}{l|lp{13.5cm}}
\toprule
\multicolumn{3}{l}{\bf Examples of the Basic EI Dataset} \\
\midrule
& \multicolumn{2}{c}{\includegraphics[height=4cm]{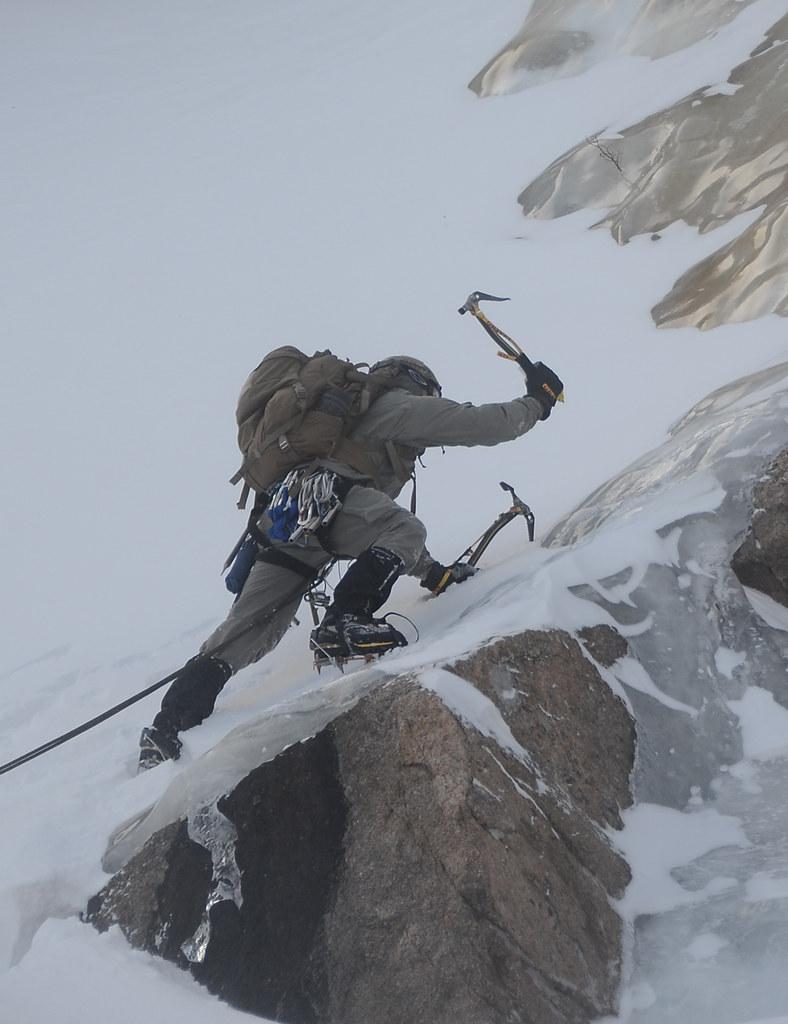}} \\
User Question & \multicolumn{2}{l}{\textit{What led to the formation of the arouse to the man in this image?}} \\
Emotional Trigger & \multicolumn{2}{p{17.5cm}}{1. Climbing a steep, snow-covered slope. 2. Physical effort and concentration. 3. Potential hazards and challenges. 4. Cold environment. 5. Determination to reach the goal.}\\
\midrule
& \multicolumn{2}{c}{\includegraphics[height=4.5cm]{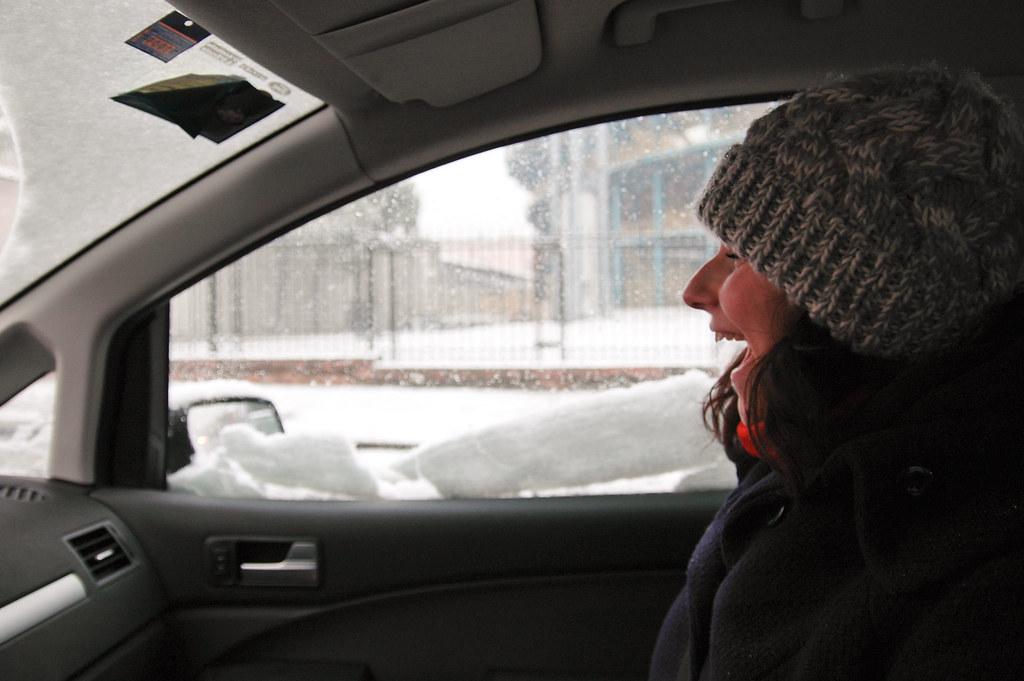}} \\
User Question & \multicolumn{2}{l}{\textit{What do you think might have caused the person's delight as they look out the window?}} \\
Emotional Trigger & \multicolumn{2}{p{17.5cm}}{1. Snowy scene outside the car. 2. Smile on her face. 3. Enjoying the view. 4. Serenity of the winter environment. 5. Excitement of experiencing a snowy day. 6. Personal or emotional connections to snowy weather or winter scenes. 7. Fresh snowfall, brightness of the snow reflecting sunlight, or peacefulness of the scene.}\\
\midrule
& \multicolumn{2}{c}{\includegraphics[height=4.5cm]{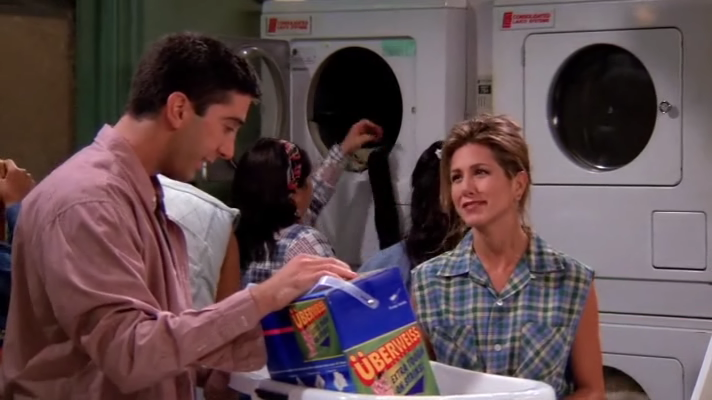}} \\
User Question & \multicolumn{2}{l}{\textit{What do you think might have caused the man holding the box in the image to become lighthearted?}} \\
Emotional Trigger & \multicolumn{2}{p{17.5cm}}{1. Holding the ``Uberweiss'' box. 2. Smiling. 3. Friendly and approachable body language. 4. Positive and relaxed atmosphere of the laundry room. 5. Interaction with others in the laundry room.}\\
\midrule
& \multicolumn{2}{c}{\includegraphics[height=4.5cm]{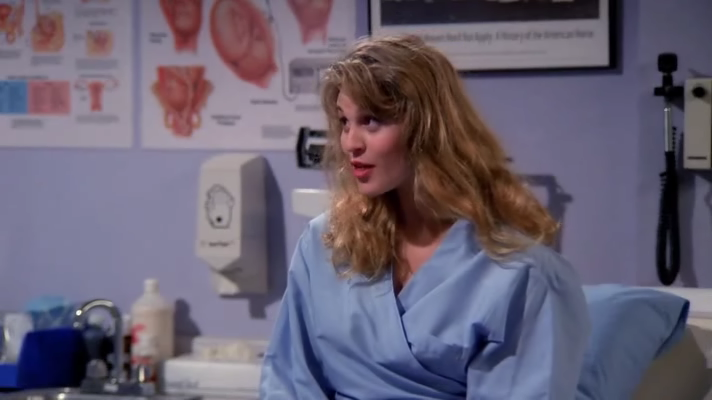}} \\
User Question & \multicolumn{2}{l}{\textit{What might have caused the woman in the image to appear content and happy?}} \\
Emotional Trigger & \multicolumn{2}{p{17.5cm}}{1. Positive news about her health. 2. Pleasant interaction with a medical professional. 3. Comforting conversation with a friend or family member. 4. Good news about her health. 5. Positive relationship with the medical staff.}\\
\midrule
& \multicolumn{2}{c}{\includegraphics[height=4.5cm]{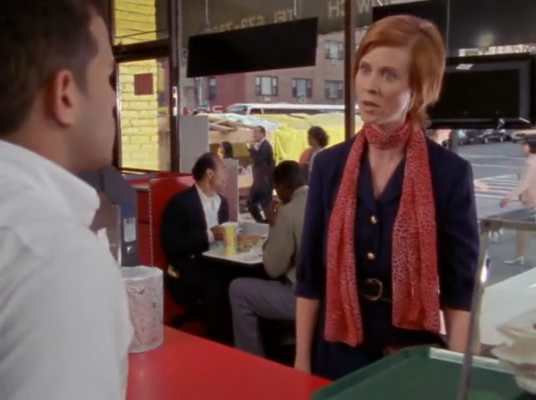}} \\
User Question & \multicolumn{2}{l}{\textit{What might have caused the woman in the image to appear irritated or angry?}} \\
Emotional Trigger & \multicolumn{2}{p{17.5cm}}{1. Service issue (mistake in order, long wait, problem with payment process). 2. Unpleasant environment (noise levels, cleanliness, presence of other customers). 3. Dissatisfaction with food or service. 4. Frustration or annoyance with the conversation or situation.}\\
\bottomrule
\end{tabular}
}
\vspace{-1mm}
\label{tab:app_basic_dataset_vis}
\end{minipage}
\end{table*}

\begin{table*}[h]
\centering
    \vspace{-2mm}
    \caption{\small Statistics of the Emotional Trigger Types (Basic Emotions).}
    \label{tab:statistics_triggers}
    \resizebox{\linewidth}{!}{
        \centering
        \begin{tabular}{cccccccccc}
        \toprule
Atmosphere & Social Interactions &  Body Movements & Facial Expressions & Objects & Performances & Outdoor Activities & Clothing & Sports & Other \\
\midrule
23.11\% & 17.17\% & 13.24\% & 9.40\% & 6.07\% & 5.06\% & 3.20\% & 3.08\% & 2.25\% & 17.41\%\\
\bottomrule

        \end{tabular}
    }
\vspace{-5mm}
\end{table*}

\begin{table*}[h]
\begin{minipage}{1\textwidth}
\centering
\vspace{-2mm}
\caption{\small Visualization of complex EI subset, an image is corresponded to multiple user questions.}
\scalebox{0.65}{
    \begin{tabular}{l|lp{13.5cm} }
\toprule
\multicolumn{3}{l}{\bf Examples of the Complex EI Subset} \\
\midrule
& \multicolumn{2}{c}{\includegraphics[height=4.5cm]{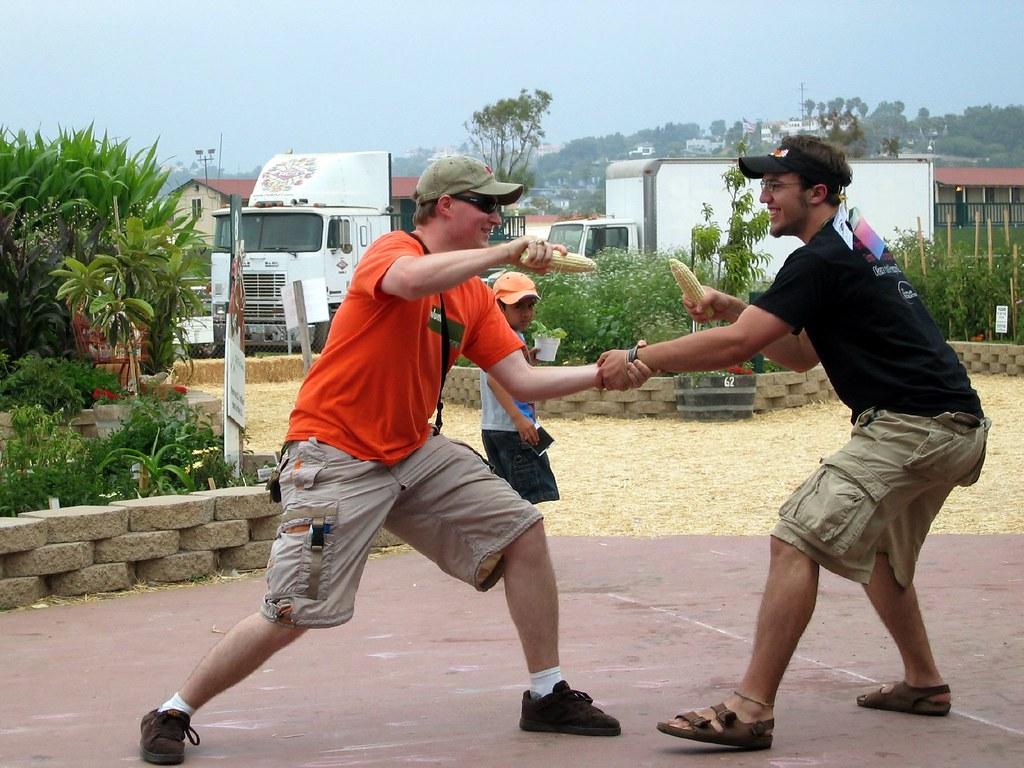}} \\
User Question (1) & \multicolumn{2}{l}{\textit{Why does the kid in the background seem excited?}} \\
Emotional Trigger & \multicolumn{2}{p{17.5cm}}{1. Head turning back. 2. Starring at the two playing with each other on the focus. 3. Sense of motion from the event. 4. Maybe excited about the desire to join them.}\\
User Question (2) & \multicolumn{2}{l}{\textit{What do you think might have caused the kid in the background of the image to be confused?}} \\
Emotional Trigger & \multicolumn{2}{p{17.5cm}}{1. Head turning back. 2. Two others acting abnormally. 3. Two others each holding a stick of corn. 4. Maybe curious about the event. 5. Maybe wondering about the motivation for the abnormality.}\\
\midrule
& \multicolumn{2}{c}{\includegraphics[height=4.5cm]{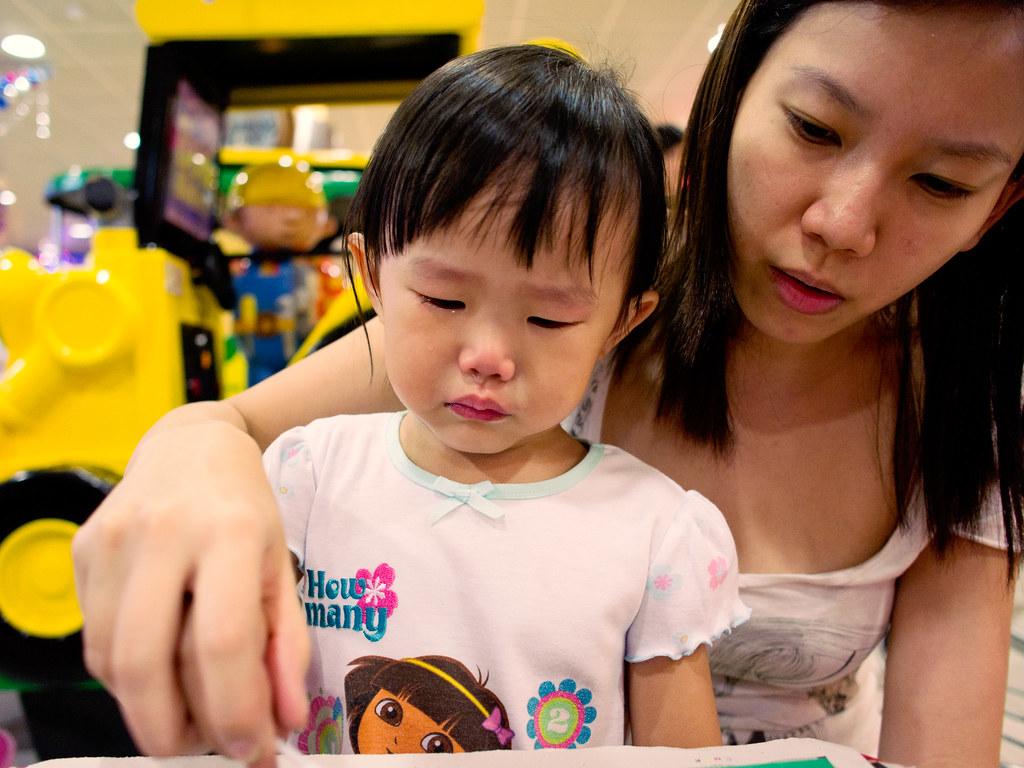}} \\
User Question (1) & \multicolumn{2}{l}{\textit{What may caused the little girl upset?}} \\
Emotional Trigger & \multicolumn{2}{p{17.5cm}}{1. Crying. 2. Can not making handiwork. 3. The woman blamed her.}\\
User Question (2) & \multicolumn{2}{l}{\textit{What may caused the little girl happy?}} \\
Emotional Trigger & \multicolumn{2}{p{17.5cm}}{1. Crying but the women comfort her. 2. Can not making handiwork. 3. Woman help her finishing the work.}\\
User Question (3) & \multicolumn{2}{l}{\textit{What may cause the woman angry?}} \\
Emotional Trigger & \multicolumn{2}{p{17.5cm}}{1. The girl is not obedient. 2. The girl can't do handiwork. 3. The girl can't learn no matter how much taught. 4. Step-by-step instruction.}\\
\midrule
& \multicolumn{2}{c}{\includegraphics[height=5cm]{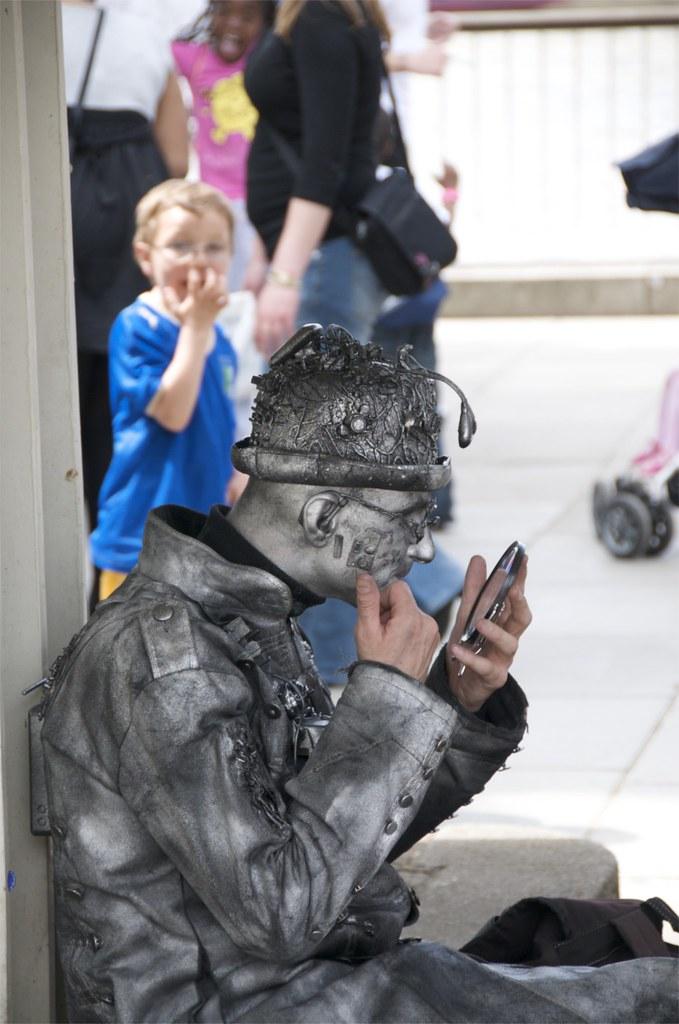}} \\
User Question (1) & \multicolumn{2}{l}{\textit{Why does the baby show the fear expression?}} \\
Emotional Trigger & \multicolumn{2}{p{17.5cm}}{1. The man's scary outfit. 2. Afraid of the man. 3. The man's makeup. 4. Covering mouth with hand.}\\
User Question (2) & \multicolumn{2}{l}{\textit{What make the baby surprise and happy?}} \\
Emotional Trigger & \multicolumn{2}{p{17.5cm}}{1. Shocking face and gesture. 2. Staring at someone. 3. Sense of unbelievable. 4. A man colored in silver on the focus. 5. Maybe shocked to see something abnormal.}\\
\midrule
& \multicolumn{2}{c}{\includegraphics[height=5cm]{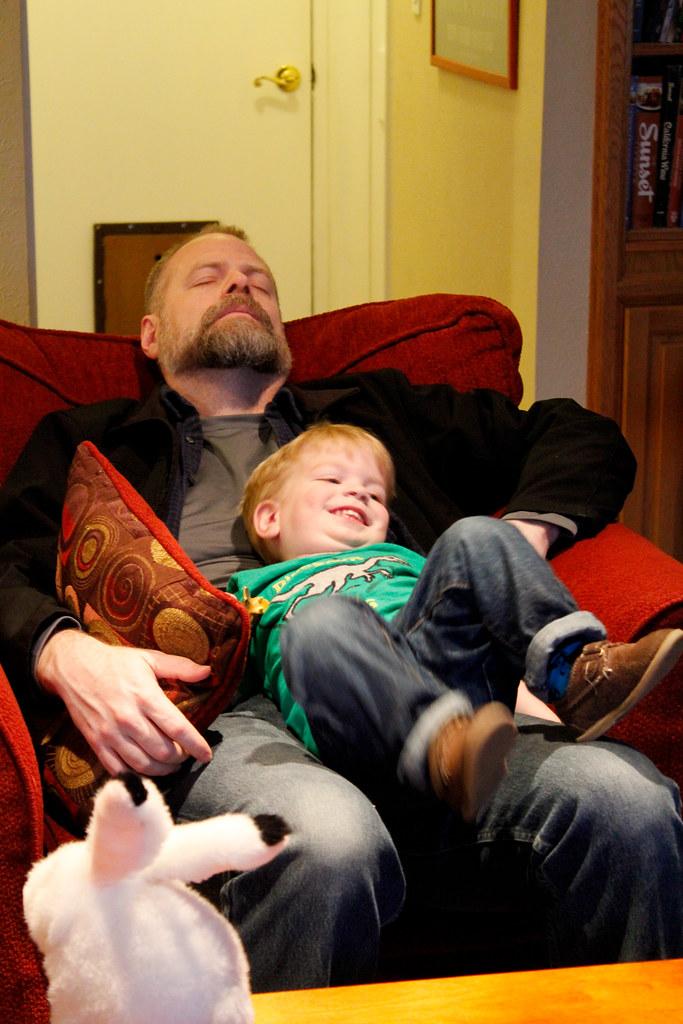}} \\
User Question (1) & \multicolumn{2}{l}{\textit{Why does this man in the picture look exhausted and annoyed?}} \\
Emotional Trigger & \multicolumn{2}{p{17.5cm}}{1. Maybe lack of Sleep. 2. Closed-eyes. 3. Taking care of a young child. 4. Tired of the child. 5. Naughty child.}\\
User Question (2) & \multicolumn{2}{l}{\textit{Why does this man being enjoyment and pleasure?}} \\
Emotional Trigger & \multicolumn{2}{p{17.5cm}}{1. Enjoying spending time with his child. 2. Child lying in arms. 3. Satisfied with the moment. 4. Sense of company of family. 5. Engaging in playful activities.}\\
\bottomrule
\end{tabular}
}
\vspace{-4mm}
\label{tab:app_com_dataset_vis}
\end{minipage}
\end{table*}

\begin{table*}[h]
\centering
    \vspace{-2mm}
    \caption{\small Statistics of the Emotional Trigger Types (Complex Emotions).}
    \label{tab:statistics_triggers_complex}
    \resizebox{\linewidth}{!}{
        \centering
        \begin{tabular}{cccccccccc}
        \toprule
Atmosphere & Social Interactions &  Body Movements & Facial Expressions & Objects & Performances & Outdoor Activities & Clothing & Sports & Other \\
\midrule
10.81\% & 23.00\% & 19.37\% & 16.22\% & 8.55\% & 0.45\% & 3.60\% & 3.60\% & 0.9\% & 13.51\%\\
\bottomrule

        \end{tabular}
    }
\vspace{-5mm}
\end{table*}

%% file: Supplementary/bench_details.tex
\begin{table*}[h]
\begin{minipage}{1\textwidth}
\centering
\vspace{-2mm}
\caption{\small Example of Hallucinations in VLLMs. Hallucinations are indicated in \textcolor{red}{red}, while other text is indicated in \textcolor{gray}{gray}.}
\scalebox{0.63}{
    \begin{tabular}{l|lp{13.5cm} }
\toprule
\multicolumn{3}{l}{\bf Examples of the Human Cleaning Process of Hallucinations} \\
\midrule
& \multicolumn{2}{c}{\includegraphics[height=4.5cm]{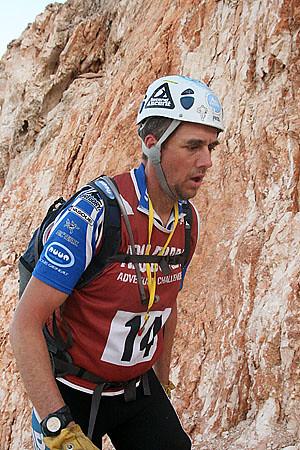}} \\
User Question & \multicolumn{2}{l}{\textit{What might have motivated the man in the image to participate in this outdoor activity, given his gear and the environment?}} \\
\midrule
Emotional Trigger (Raw) & \multicolumn{2}{p{17.5cm}}{\textcolor{gray}{1. Determination and concentration. 2. Challenge of the race or trail. 3. Personal goals. 4. Desire to improve mountain biking skills. 5. Well-prepared gear. 6. Environmental factors (rocky slope, weather conditions).} \textcolor{red}{7. Doing mountain biking}.}\\
\midrule
& \multicolumn{2}{c}{\includegraphics[height=4.5cm]{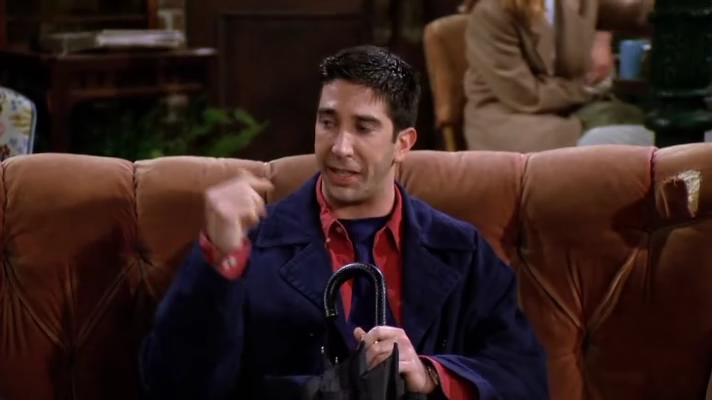}} \\
User Question & \multicolumn{2}{l}{\textit{What could have caused the man in the image to appear outraged or hostile?}} \\
\midrule
Emotional Trigger (Raw) & \multicolumn{2}{p{17.5cm}}{\textcolor{red}{1. Holding a black bag}. \textcolor{gray}{2. Animated conversation or gesture. 3. Furrowed eyebrows. 4. Open mouth. 5. Wide or squinting eyes. 6. Leaning forward or gesturing with hands.} \textcolor{red}{7. Brown couch (as a place where he typically relaxes or discusses matters)}}\\
\midrule
& \multicolumn{2}{c}{\includegraphics[height=4.5cm]{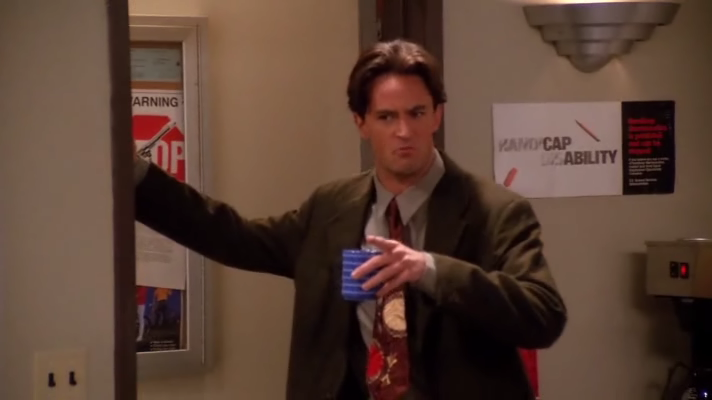}} \\
User Question & \multicolumn{2}{l}{\textit{What might have caused the man in the image to be angry or upset?}} \\
\midrule
Emotional Trigger (Raw) & \multicolumn{2}{p{17.5cm}}{\textcolor{red}{1. KANO CAP ABILITY sign on the wall. 2. Feeling overwhelmed or pressured by his workload. 3. Undervalued or overworked in his professional role} \textcolor{gray}{4. Recent events or interactions in the workplace that have caused stress or frustration. 5. Tension and stress in his body language (posture, grip on the mug)}.}\\
\midrule
& \multicolumn{2}{c}{\includegraphics[height=4.5cm]{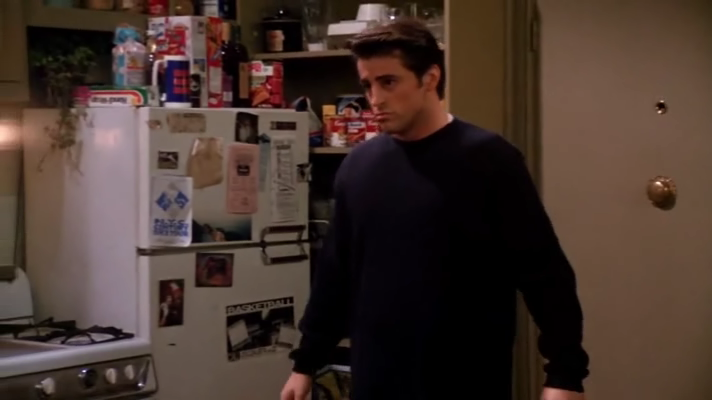}} \\
User Question & \multicolumn{2}{l}{\textit{What might have caused the man in the image to appear angry or frustrated?}} \\
\midrule
Emotional Trigger (Raw) & \multicolumn{2}{p{17.5cm}}{\textcolor{red}{1. Disagreement with a family member. 2. Concern about a meal he is preparing. 3. Problem at work that he is thinking about while in the kitchen.} \textcolor{gray}{4. Serious or intense mood due to work-related issue or concern.}}\\
\bottomrule
\end{tabular}
}
\vspace{-4mm}
\label{tab:hullucination}
\end{minipage}
\end{table*}

\begin{table*}[h]
\begin{minipage}{1\textwidth}
\centering
\vspace{-2mm}
\caption{\small The Human in the Loop process instills Commonsense Knowledge into the dataset. Text \textcolor{orange}{orange} represents added commonsense knowledge.}
\scalebox{0.65}{
    \begin{tabular}{l|lp{13.5cm} }
\toprule
\multicolumn{3}{l}{\bf Examples of Data Cleaning for Commonsense Knowledge} \\
\midrule
& \multicolumn{2}{c}{\includegraphics[height=4.5cm]{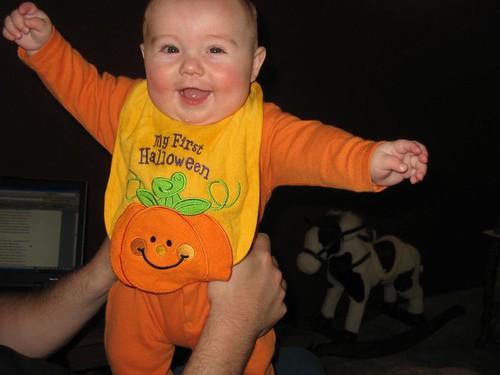}} \\
User Question & \multicolumn{2}{l}{\textit{What might have caused the baby's delight in this image?}} \\
\midrule
Emotional Trigger & \multicolumn{2}{p{17.5cm}}{\textcolor{orange}{1. Halloween costume and bib with a pumpkin design.} 2. Interaction with the person holding them up. 3. Festive atmosphere and attention from the person holding them up. \textcolor{orange}{4. First Halloween experience}.}\\
\midrule
& \multicolumn{2}{c}{\includegraphics[height=4.5cm]{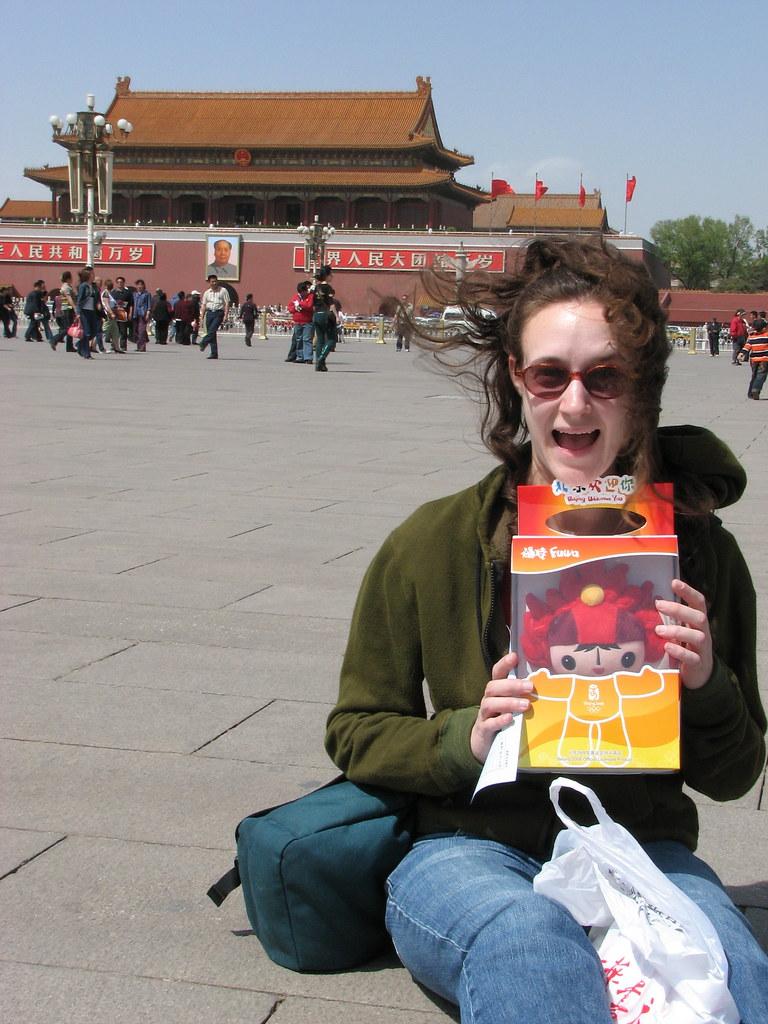}} \\
User Question & \multicolumn{2}{l}{\textit{What led to the excitement on the woman's face?}} \\
\midrule
Emotional Trigger & \multicolumn{2}{p{17.5cm}}{\textcolor{orange}{1. A toy written ``Beijing Welcome''. 2. Taking a photo with Tienanmen Square. 3. First time to Beijing}.}\\
\midrule
& \multicolumn{2}{c}{\includegraphics[height=4.5cm]{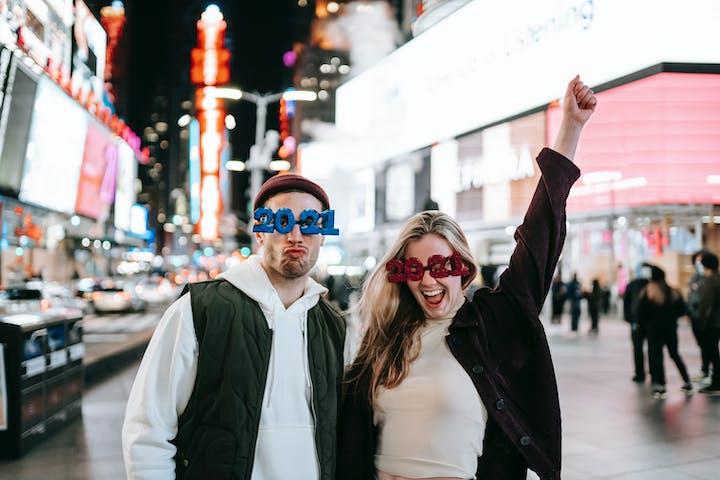}} \\
User Question & \multicolumn{2}{l}{\textit{What might have caused the man in the image to become excited and make a funny face?}} \\
\midrule
Emotional Trigger & \multicolumn{2}{p{17.5cm}}{\textcolor{orange}{1. Celebratory event or milestone related to the year 2021.} 2. Excitement and joy. 3. Playful or lighthearted moment shared between the man and the woman. 4. Achievement or personal milestone. 5. Festive and celebratory atmosphere.}\\
\midrule
& \multicolumn{2}{c}{\includegraphics[height=4.5cm]{img/appendix/excitement_11293.jpg}} \\
User Question & \multicolumn{2}{l}{\textit{Why does the kid in the background seem excited?}} \\
\midrule
Emotional Trigger & \multicolumn{2}{p{17.5cm}}{1. Head turning back. 2. Starring at the two playing with each other on the focus. 3. Sense of motion from the event. \textcolor{orange}{4. Maybe excited about the desire to join them.}}\\
\bottomrule
\end{tabular}
}
\vspace{-4mm}
\label{tab:common_sense}
\end{minipage}
\end{table*}